\documentclass[authoryear,review,12pt]{elsarticle}
\usepackage{lineno}
\usepackage{epsfig}
\usepackage{subcaption}
\usepackage{multirow}%
\usepackage{multicol}
\usepackage{amsmath} % Enhanced mathematical features
\usepackage{mathtools} % Extension of amsmath
\usepackage{amssymb} % Additional mathematical symbols
\usepackage{stmaryrd} % Additional mathematical symbols
\usepackage[title]{appendix}%
\usepackage{xcolor}%
\usepackage{textcomp}%
\usepackage{manyfoot}%
\usepackage{algorithm}%
\usepackage{algpseudocode}%
\usepackage{listings}%

\usepackage{printlen}
\usepackage{layouts}
\usepackage{hyperref}
\usepackage{cleveref}
\usepackage{makecell}
\usepackage{xfrac}
\usepackage{pgfplots, pgf}
\usepackage{booktabs}
\usepackage{adjustbox}
\usepackage{siunitx}
\usepackage{xargs}
\usepackage[colorinlistoftodos,prependcaption]{todonotes}
\usepackage{csquotes}
\usepackage{acronym}
\usepackage{xspace}
\usepackage{comment}
\usepackage{placeins} % Include this package in your preamble
\usepackage{seqsplit}
\usepackage{rotating}
\usepackage{breqn}
\usepackage{floatflt}
\usepackage{microtype}
\usepackage{epstopdf}

\newcommand{\textttnew}[1]{\texttt{\seqsplit{#1}}}

\newcommand{\exps}[1]{\mathbf{e}^{(#1)}}
\renewcommand{\eqref}[1]{(\ref{#1})}
\renewcommand{\exp}[1]{\text{exp}{\left(#1\right)}}
\renewcommand{\vec}[1]{\boldsymbol{#1}}

\newcommand{\set}[1]{\{#1\}}
\newcommand{\bemph}[1]{\textbf{\emph{#1}}}
\newcommand{\card}[1]{\lvert #1 \rvert}

\newcommand{\indic}[1]{\llbracket #1 \rrbracket}

\newcommand{\floor}[1]{\lfloor #1 \rfloor}

\newcommand{\expn}{\mathrm{e}}
\newcommand{\expnumber}[2]{{#1}\expn{#2}}
\newcommand{\given}[2]{{#1} \, |\, {#2}}
\newcommand{\givens}[2]{{#1} | {#2}}
\newcommand{\fromto}{\rightarrow}
\newcommand*{\from}{\colon}
\newcommand{\coloneq}{\!\!:=}
\newcommand{\product}{\cdot}

\newcommand{\IR}{\mathbb{R}}
\newcommand{\IN}{\mathbb{N}}

\newcommand{\IE}{\mathbb{E}}
\newcommand{\cY}{\mathcal{Y}}
\newcommand{\cX}{\mathcal{X}}
\newcommand{\cD}{\mathcal{D}}
\newcommand{\hy}{\hat{y}}
\newcommand{\vhy}{\vec{\hat{y}}}
\newcommand{\vy}{\vec{y}}
\newcommand{\zo}{0\text{-}1}
\newcommand{\nc}{C}
\newcommand{\ncs}{c}
\newcommand{\holm}{\tau}

\newcommand{\setp}{\mathbb{P}}
\newcommand{\loss}{\ell}

\newcommand{\ild}{L}
\newcommand{\hild}{\hat{L}}
\newcommand{\data}{\mathcal{L}}
\newcommand{\mi}{I}
\newcommand{\hmi}{\widehat{I}}
\newcommand{\ent}{H}
\newcommand{\gentype}{\mathsf{g}_{r}}

\newcommand{\noise}{\epsilon}

\newcommand{\prob}{p}
\newcommand{\hprob}{\hat{\prob}}
\newcommand{\pgiven}[2]{\prob_{\givens{#1}{#2}}}
\newcommand{\hpgiven}[2]{\hprob_{\givens{#1}{#2}}}
\newcommand{\pjoint}{\prob_{(X, Y)}}
\newcommand{\hpjoint}{\hprob_{(X, Y)}}
\newcommand{\pygivenx}{\pgiven{Y}{X}}
\newcommand{\hpygivenx}{\hpgiven{Y}{X}}
\newcommand{\pxgiveny}{\pgiven{X}{Y}}
\newcommand{\hpxgiveny}{\hpgiven{X}{Y}}
\newcommand{\pxmarg}{\prob_{X}}
\newcommand{\hpxmarg}{\hprob_{X}}
\newcommand{\pymarg}{\prob_{Y}}
\newcommand{\hpymarg}{\hprob_{Y}}
\newcommand{\mvn}{\textsc{MVN}}

\newcommand{\cM}{\mathcal{M}}

\newcommand{\cH}{\mathcal{H}}
\newcommand{\cF}{\mathcal{F}}
\newcommand{\CM}[2]{\boldsymbol{M}_{#1}^{#2}}
\newcommand{\acc}{\vec{a}}

\newcommand{\sixpt}{\fontsize{6.0}{7.2}\selectfont}
\newcommand{\on}{\operatorname}
\newcommand{\textscsix}[1]{\textsc{\sixpt{#1}}}

\newcommand{\met}{\on{m}}
\newcommand{\metric}[1]{\met_{\textscsix{#1}}}
\newcommand{\metricbp}[2]{\metric{#1}{(#2)}}
\newcommand{\tp}{\metric{tp}}
\newcommand{\tn}{\metric{tn}}
\newcommand{\fp}{\metric{fp}}
\newcommand{\fn}{\metric{fn}}

\newcommand{\argmin}{\on*{\arg\,\min}}
\newcommand{\argmax}{\on*{\arg\,\max}}
\newcommand{\inte}{\int\limits}

\newcommand{\err}{Error-rate\xspace}
\newcommand{\fpr}{FPR\xspace}
\newcommand{\fnr}{FNR\xspace}

\newacro{MCC}{Matthews correlation coefficient}
\newacro{BER}{balanced error rate}
\newacro{FPR}{false positive rate}
\newacro{FNR}{false negative rate}
\newacro{CM}{confusion matrix}
\acrodefplural{CM}[CMs]{confusion matrices}
\newcommand{\accuracy}{Accuracy\xspace}

\newcommand{\tabpfn}{TabPFN\xspace}
\newcommand{\autogluon}{AutoGluon\xspace}
\newcommand{\fetmean}{\textsc{FET-Mean}\xspace}
\newcommand{\fetmedian}{\textsc{FET-Median}\xspace}
\newcommand{\pttmc}{\textsc{PTT-Majority}\xspace}

\newcommand{\midpoint}{\textsc{Mid-Point}\xspace}
\newcommand{\llmis}{log-loss\xspace}
\newcommand{\llmi}{\textsc{Log-Loss}\xspace}
\newcommand{\calmi}{\textsc{Cal Log-Loss}\xspace}
\newcommand{\ircalmi}{\textsc{IR Cal Log-Loss}\xspace}
\newcommand{\pscalmi}{\textsc{PS Cal Log-Loss}\xspace}
\newcommand{\betacalmi}{\textsc{Beta Cal Log-Loss}\xspace}
\newcommand{\tscalmi}{\textsc{TS Cal Log-Loss}\xspace}
\newcommand{\hbcalmi}{\textsc{HB Cal Log-Loss}\xspace}
\newcommand{\pcsoftmax}{\textsc{PC-Softmax}\xspace}
\newacro{PC-softmax}{probability-corrected softmax}

\newcommand{\makename}[3][s]{%
	\expandafter\newcommand\csname #2\endcsname{#3\xspace}%
	\expandafter\newcommand\csname #2s\endcsname{#3#1\xspace}%
}

\makename{genmethod}{generation method}
\makename{cgenmethod}{Generation method}
\makename{ite}{information theory}
\makename{slt}{statistical learning theory}
\makename{bc}{binary classifier}
\makename{bca}{binary classification algorithm}
\makename{bpmc}{marginal Bayes predictor}
\makename{mc}{marginal classifier}
\makename{bp}{Bayes predictor}
\makename{rg}{random guessing}
\makename{ildd}{IL-Dataset}
\makename{dmlp}{deep multi-layer perceptron}
\makename{errn}{error rate}
\makename{berr}{Bayes error rate}
\makename{relu}{\texttt{ReLU}}
\makename{linear}{\texttt{Linear}}
\makename{sgd}{\texttt{SGD}}
\makename{adam}{\texttt{Adam}}
\makename{rmsprop}{\texttt{RMSProp}}
\makename{skopt}{\texttt{scikit-optimize}}
\makename{infoselect}{\texttt{InfoSelect}}
\makename{netcal}{\texttt{netcal}}
\makename{pytorch}{\texttt{PyTorch}}
\makename{keras}{\texttt{Keras}}
\makename{scipy}{\texttt{scipy}}
\makename{openml}{\texttt{openml}}
\makename{fastai}{\texttt{fastai}}

\newacro{ML}{machine learning}
\newacro{MLA}{machine learning algorithm}
\newacro{DL}{deep learning}
\newacro{MI}{mutual information}
\newacro{IL}{information leakage}
\newacro{ILD}{information leakage detection}
\newacro{ILQ}{information leakage quantification}
\newacro{LAS}{leakage assessment score}

\newacro{IoT}{Internet of Things}
\newacro{KL}{Kullback-Leibler}
\newacro{CKE}[\textttnew{CKE}]{\textttnew{ClientKeyExchange}}
\newacro{FIN}[\textttnew{FIN}]{\textttnew{Finished}}
\newacro{CCS}[\textttnew{CCS}]{\textttnew{ChangeCipherSpec}}
\newacro{PMS}[\textttnew{PMS}]{\emph{pre-master secret}}
\newacro{PS}[\textttnew{PS}]{padding string}
\newacro{CCE}{categorical cross-entropy}

\newacro{AutoML}{automated machine learning}
\newacro{CASH}{combined algorithm selection and hyperparameter optimization}
\newacro{PDF}{probability density function}
\newacro{PMF}{probability mass function}
\newacro{SCA}{side-channel attack}
\newacro{MAE}{mean absolute error}
\newacro{NMAE}{normalized mean absolute error}
\newacro{SE}{standard error}
\newacro{SUT}{system under test}

\newacro{KFCV}{$K$-fold cross-validation}
\newacro{MCCV}{Monte Carlo cross-validation}
\newacro{MVN}{multivariate normal}
\newacro{MLP}{multi-layer perceptron}
\newacro{NN}{neural network}
\newacro{XT}{extra trees classifier}
\newacro{RF}{random forest classifier}
\newacro{GBM}{gradient boosting machine}
\newacro{LightGBM}{light gradient boosting machine}
\newacro{CatBoost}{categorical boosting machine}
\newacro{XGBoost}{eXtreme gradient boosting machine}
\newacro{OTT}{one-sample t-test}
\newacro{FET}{Fisher's exact test}
\newacro{PTT}{paired t-test}
\newacro{GMM}{Gaussian mixture model}
\newacro{HDDC}{high-dimensional data clustering}
\newacro{EM}{expectation-maximization}
\newacro{AIC}{Akaike information criterion}
\newacro{MINE}{mutual information neural estimation}
\newacro{HPO}{hyperparameter optimization}
\newacro{MSE}{mean squared error}
\newacro{HB}{Hyperband}
\newacro{BO}{Bayesian optimization}
\newacro{BOHB}{Bayesian optimization and Hyperband}

\makeatletter
\newcommand\thefontsize{The current font (\f@family ) size is: \f@size pt}
\makeatother
\journal{Information Sciences}

\begin{document}
\begin{frontmatter}

\title{Information Leakage Detection through Approximate Bayes-optimal Prediction}

\author[upbaddress,rub]{Pritha Gupta\corref{corauthor}}
\ead{prithag@mail.upb.de}
\author[luhai]{Marcel Wever}
\ead{marcel.wever@ai.uni-hannover.de}
\author[lmu,mcml,dfki]{Eyke Hüllermeier}
\ead{eyke@lmu.de}

\affiliation[upbaddress]{organization={Software Innovation Lab, Paderborn University}, 
% addressline={Department of Computer Science}, 
city={Paderborn}, 
country={Germany}}

\affiliation[rub]{organization={Research Center for Trustworthy Data Science and Security, Ruhr University Bochum, Germany}, 
% addressline={Department of Computer Science}, 
city={Bochum}, 
country={Germany}}

\affiliation[luhai]{organization={L3S Research Center, Leibniz University Hannover}, 
city={Hannover}, 
country={Germany}}

\affiliation[lmu]{organization={Institute of Informatics, University of Munich}, 
%addressline={}, 
city={Munich}, 
country={Germany}}

\affiliation[mcml]{organization={Munich Center for Machine Learning (MCML)}, 
%addressline={}, 
city={Munich}, 
country={Germany}}

\affiliation[dfki]{organization={German Centre for Artificial Intelligence (DFKI/DSA)}, 
%addressline={}, 
city={Kaiserslautern}, 
country={Germany}}

\cortext[corauthor]{Corresponding author}

\begin{abstract}
In today's data-driven world, the proliferation of publicly available information raises security concerns due to the \ac{IL} problem.
\ac{IL} involves unintentionally exposing sensitive information to unauthorized parties via observable system information.
Conventional statistical approaches rely on estimating \ac{MI} between observable and secret information for detecting \acp{IL}, face challenges of the curse of dimensionality, convergence, computational complexity, and \ac{MI} misestimation.
Though effective, emerging supervised \acl{ML} based approaches to detect \acp{IL} are limited to binary system sensitive information and lack a comprehensive framework.
To address these limitations, we establish a theoretical framework using \slt and \ite to quantify and detect \ac{IL} accurately. 
Using \acl{AutoML}, we demonstrate that \ac{MI} can be accurately estimated by approximating the typically unknown \bp's \llmi and accuracy.
Based on this, we show how \ac{MI} can effectively be estimated to detect \acp{IL}.
Our method performs superior to state-of-the-art baselines in an empirical study considering synthetic and real-world OpenSSL TLS server datasets.
 %We compare our \ac{MI} estimation methods against current baselines on synthetic datasets generated using a \acl{MVN} distribution with known \ac{MI}.
 %We introduce a cut-off technique using a one-sided statistical test to detect \ac{IL} and employ the Holm-Bonferroni correction to increase confidence in detection decisions.
 %Our study evaluates \acl{ILD} approaches on real-world OpenSSL TLS Server data, and highlights the superiority of \ac{MI} estimation using \llmis in accurately detecting \acp{IL} and estimating \ac{MI} in synthetic datasets.
\end{abstract}

%\includegraphics[width=\textwidth]{grabs}

\begin{keyword}
Information Leakage Detection, Mutual Information, Bayes-optimal Predictor, AutoML, Statistical Tests, Privacy
\end{keyword}

\end{frontmatter}
	

\section{Introduction}
\label{sec:introduction}
The rapid proliferation of publicly available data, coupled with the increasing use of \ac{IoT} technologies in today's data-driven world, has magnified the challenge of \ac{IL}, posing substantial risks to system security and confidentiality \citep{john2002compression}.
\ac{IL} occurs when sensitive or confidential information is inadvertently exposed to unauthorized individuals through observable system information \citep{hettwer_application_2020}.
This problem can lead to severe consequences, ranging from potential electrical blackouts to the theft of critical information like medical records and military secrets, making the efficient detection and quantification of \ac{IL} of paramount importance \citep{hettwer_application_2020,shabtai2012dl}.

According to \ite, quantifying \ac{IL} typically involves estimating \ac{MI} between observable and secret information \citep{chatzikokolakis2010statistical}.
Despite being a pivotal measure, \ac{MI} is difficult to compute for high-dimensional data, facing challenges such as the \emph{curse of dimensionality}, convergence, and computational complexity \citep{gao2015misurvey}. 
Traditional statistical estimation methods often struggle with all of these challenges, while more recent robust non-parametric approaches with improved convergence rates still find high-dimensional scenarios challenging \citep{moon2021ensemble}.

In recent years, \ac{ML} techniques have gained popularity in \ac{ILD}, particularly for performing \acp{SCA} on cryptographic systems \citep{picek2023sok}.
These systems release the \emph{observable information} via many modes called the side-channels, such as network messages, CPU caches, power consumption, or electromagnetic radiation, which are exploited by \acp{SCA} to reveal secret inputs (secret keys, plaintexts), potentially rendering cryptographic protections ineffective \citep{dlla2021moos, hettwer_application_2020}. 
Therefore, detecting the existence of a side-channel is equivalent to uncovering \ac{IL} \citep{hettwer_application_2020}.
In this field, the most relevant literature uses \ac{ML} to perform \acp{SCA} rather than preventing side-channels through early detection of \acp{IL} \citep{hettwer_application_2020}.
Current \ac{ML}-based methods in this realm detect side-channels to prevent \acp{SCA} and protect the system on both algorithmic and hardware levels \citep{dlla2021moos}.
These approaches leverage observable information to classify systems as vulnerable (with \ac{IL}) or non-vulnerable (without \ac{IL}) \citep{Perianin2021}. 
They extract observable information from secure systems, categorizing them as non-vulnerable (labeled $0$), then introduce known \acp{IL} to categorize them as vulnerable (labeled $1$), creating a classification dataset for the learning model.
However, this approach is limited to domain-specific scenarios and cannot be easily transferred to detect other unknown leakages \citep{Perianin2021}.

Recent promising \ac{ML}-based methods proposed for estimating \ac{MI} within classification datasets grapple with challenges related to convergence and computational complexity \citep{cristiani2020mineild}, and others may underestimate \ac{MI} or miss specific subclasses of \ac{IL} \citep{zhenyue2019rethinking}.
Recent advancements have demonstrated the effectiveness of \ac{ML}-based techniques in directly detecting \ac{IL} by analyzing the accuracy of the supervised learning models on extracted system data \citep{dlla2021moos}. 
Yet, these methods exhibit limitations in handling imbalanced and noisy real-world datasets, commonly encountered in practical scenarios, and tend to miss \acp{IL} by producing false negatives \citep{jiajia_novel_2020, curse2018picek}. 

To address these limitations, in our prior work, we proposed utilizing \bcs integrated with \ac{FET} and \ac{PTT} statistical tests to account for imbalance \citep{ild2022pritha}.
Despite its merits, this approach is limited to binary classification tasks and needs a comprehensive theoretical framework.

\paragraph{Our Contributions}
\label{sec:contributions}
We address these shortcomings with the following contributions:
\begin{itemize}\setlength\itemsep{0em}
    \item A novel \ac{ILD} framework with a generalized \acs{LAS} metric to assess \acp{IL} accurately.
    \item \ac{MI} estimators approximating Bayes' performance via automated machine learning. 
    \item Bayes’ performance in terms of \llmi is used to address the data imbalance.
    \item \acs{ILD} by thresholding on \acs{MI} with Holm-Bonferroni correction for robust detection.
    \item Empirical study of \ac{ILD} approaches vs. baselines to counter Bleichenbacher \ac{SCA}.

\end{itemize}

\section{Information Leakage Detection Problem}
\label{sec:ilproblem}
In this section, we formalize the \ac{ILD} task of categorizing a system as vulnerable or non-vulnerable using the proposed generalized \ac{LAS} measure to quantify \ac{IL}, subsequently used to detect \ac{IL} in the system.
\ac{LAS} is evaluated by comparing the (approximate) performance of the \bp and \bpmc; when using \llmis, it reduces to \ac{MI}, a standard measure for \ac{IL} quantification.
We also briefly introduce the concepts of \bp and \ac{MI}, with details of \slt using the notations defined in \Cref{tab:notations}. 

\subsection{Formal Setting}
\label{sec:ilproblem:formalism}
\ac{ILD} aims to identify unintended disclosure of \emph{secret information} through \emph{observable information} of the system. 
The \ac{ILD} algorithm analyzes the system dataset $\cD = \set{(\vec{x}_i, y_i)}_{i=1}^N \subset \cX \times \cY, N \in \IN$, where $\cX = \IR^d$ represents observable information and $\cY = [\nc]$ represents secret information as categorical classes.
The goal is to label $\cD$ with $1$ indicating \ac{IL} and $0$ its absence, represented by the mapping $\ild$ as
\begin{equation*}
\ild: \bigcup_{N \in \IN} (\cX \times \cY)^N \fromto \set{0,1} \, ,
%\label{eq:ild}
\end{equation*}
which takes a dataset $\cD$ of any size as input and outputs the decision on the presence of \ac{IL} in the system. 
The \ac{ILD} approach produces the mapping $\hild$ and predicts \acp{IL} in the given system.

Let $\data = \set{(\cD_i, z_i)}_{i=1}^{N_L}$ be an \textbf{\ildd}, such that ${N_L} \in \IN, z_i \in \set{0,1}$ and $\vec{z}=(z_1, \ldots, z_{N_L})$ be the ground truth vector generated by $\ild$.
The predicted \acp{IL} produced by $\hild$ are denoted as the vector $\vec{\hat{z}}=(\hat{z}_1, \ldots, \hat{z}_{N_L})$, such that $\hat{z}_i = \hild(\cD_i)$.
The performance of an \ac{ILD} approach ($\hild$) is measured using standard classification metrics ($\met_{(\cdot)}(\vec{z}, \vec{\hat{z}})$) (c.f. \ref{asec:impdetails:evaluationmetric}).

\subsection{Fundamentals}
\label{sec:fundamentals}
We briefly introduce the concepts of \bp and \ac{MI}, including the details of the classification problem.

\subsubsection{Mutual Information}
\label{sec:fundamentals:mi}
\Acf{MI} measures the extent to which knowledge of one random variable informs about another, quantifying their dependence degree \citep{cover2006elements}. 
Consider a pair of random variables $X$ and $Y$ with joint distribution $\pjoint(\cdot)$ on $\cX \times \cY$. 
We assume that $X$ is a continuous $d$-dimensional real-valued random variable ($\cX = \IR^d$), and $Y$ is a discrete random variable with $\nc$ possible values---as in the \ac{IL} scenario relevant to us \citep{chatzikokolakis2010statistical}.
Let the measure $P$ induces a marginal \ac{PDF} on $\cX$ and $\cY$ denoted by $\pxmarg(\cdot)$ and $\pymarg(\cdot)$ of the joint distribution $\pjoint(\cdot)$, which induces the conditional distributions $\pygivenx(\cdot)$ and $\pxgiveny(\cdot)$. 

The entropy of a discrete random variable $Y$ is defined as
\begin{equation}
\ent(Y) = -\sum_{y \in \cY} \pymarg(y) \log (\pymarg(y)) \, ,
\label{eq:enty}
\end{equation}
where $0 \log(0) = 0$ by definition.
It reaches the maximum value of $\log(\nc)$ when outcomes are equally likely ($\pymarg(y) = 1/\nc, \forall y \in [\nc]$), indicating complete uncertainty of the outcome.
The minimum value of $0$ occurs in the case of a Dirac measure when only one outcome is certain, i.e., $\pymarg(y) = 0, \forall y \in [\nc] \setminus {\ncs}, \, \pymarg(\ncs) = 1$.

The conditional entropy of $Y$ given $X$ is defined as
\begin{equation*}
\ent(\given{Y}{X}) = -\inte_{\vec{x} \in \cX} \pxmarg(\vec{x}) \sum_{y \in \cY} \pygivenx(\given{y}{\vec{x}}) \log(\pygivenx(\given{y}{\vec{x}})) \, \mathrm{d}\vec{x}\, . 
%\label{eq:cond_entygivenx}
\end{equation*}
Conditional entropy measures the residual uncertainty in one random variable $Y$ given knowledge of the other $X$\,---\ more specifically, it measures the \emph{expected} residual uncertainty, with the expectation taken with respect to the marginal distribution of $Y$.
It reaches its maximum value of $\ent(Y)$, when $X$ does not inform about $Y$, i.e., $\ent(\given{Y}{\vec{x}}) = \pymarg(\cdot), \forall \vec{x} \in \cX$, and the minimum entropy of $0$ occurs when $X$ completely determines $Y$ and again all conditionals $\pygivenx(\given{\cdot}{\vec{x}})$ are Dirac distributions.

\ac{MI} measures the reduction of uncertainty about variable $Y$ by observing variable $X$:
\begin{equation}
\mi(X; Y) = \ent(Y)-\ent(\given{Y}{X}) \, .
\label{eq:mi_y}
\end{equation}
Plugging in the expressions for (conditional) entropy and rearranging terms shows that \ac{MI} equals the \ac{KL} divergence of the joint distribution $\pjoint(\cdot)$ from the product of the marginals (i.e., the joint distribution under the assumption of independence):
\begin{align}
\mi(X;Y) & = \ent(Y) - \ent(\given{Y}{X}) \nonumber \\
&= \inte_{\vec{x} \in \cX}\sum_{y \in \cY}{\pjoint(\vec{x}, y) \log {\left({\frac {\pjoint(\vec{x}, y)}{\pxmarg(\vec{x}) \product \pymarg(y)}}\right)}} \, d\vec{x} \, .
\label{eq:mi_integral}
\end{align}
\ac{MI} is a symmetric measure ranging from $0$ to $\min(\set{\ent(X), \ent(Y)})$, where $0$ indicates independence and the maximum value signifies full dependence~\citep{cover2006elements}. 
In this paper, \ac{MI} is measured in \bemph{bits} using base-2 logarithms ($\log(\cdot)$).

\subsubsection{Classification Problem}
\label{sec:fundamentals:classification}
In the realm of classification, the learning algorithm (learner) is provided with a training dataset $\cD = \set{(\vec{x}_i, y_i)}_{i=1}^N \subset \cX \times \cY$ of size $N \in \IN$, where $\cX = \IR^d$ is the input (instance) space and $\cY = \set{1, 2, \ldots, \nc} = [\nc], \, \nc \in \IN$ the output (categorical classes) space \citep{vapnik1991principles}, and the $(\vec{x}_i, y_i)$ are assumed to be independent and identically distributed (i.i.d.) according to $\pjoint(\cdot)$.
The primary goal of the learner in standard classification is to induce a hypothesis $h \from \cX \fromto \cY, \, h \in \cH$, with low generalization error (risk)
\begin{equation}
R(h) = \IE[\loss(y, h(\vec{x}))] = \inte_{\cX \times \cY} \loss(y, h(\vec{x})) \, d \, \pjoint(\vec{x}, y) \, ,
\label{eq:risk}
\end{equation}
where $\cH$ is the underlying hypothesis space, $\loss \from \cY \times \cY \fromto \IR$ is a loss function, and $\pjoint(\cdot)$ is the joint probability measure modeling the underlying data-generating process.
The risk minimizer defined as $h^* = \argmin_{h \in \cH} R(h)$, achieves the minimum expected loss $\IE[\loss(\cdot)]$ in terms of the loss function $\loss$ across the entire joint distribution $\pjoint(\cdot)$.
The $\zo$ loss defined as $\loss_{01}(y,\hy) \coloneq \indic{\hy \neq y}$, is commonly used in standard classification.
%The measure $\pjoint(\cdot)$ in \eqref{eq:risk} induces marginal probability (density or mass) functions on $\cX$ and $\cY$ as well as a conditional probability of the class $Y$ given an input instance $\vec{x}$, i.e., $\pjoint(\vec{x}, y) = \pygivenx(\given{y}{\vec{x}}) \times \pxmarg(\vec{x})$.
In practice, the $\pjoint(\vec{x}, y)$ is not directly observed by the learner, so minimizing the risk \eqref{eq:risk} is not feasible.
Instead, learning in a standard classification setting is commonly accomplished by minimizing (a regularized version of) \emph{empirical risk} for $h$:
\begin{equation}
R_{\text{emp}}(h) = \dfrac{1}{N}\sum_{i=1}^N \loss(y_i , h(\vec{x}_i)) \, .
\label{eq:erisk}
\end{equation}
In the subsequent discussions, we denote by $g = \argmin_{h \in \cH} R_{\text{emp}}(h)$ the (learned) hypothesis that minimizes \eqref{eq:erisk}, i.e., the empirical risk minimizer, which in principle is the best possible approximation (empirical estimation) $h^*$ and is an approximation thereof \citep{vapnik1991principles}.
%In practice, for the available finite sampled dataset $\cD$, $g$ is the best possible approximation (empirical estimation) of the \emph{true} risk minimizer $h^*$ and is an approximation thereof. 

\paragraph{Probabilistic Classification}
\label{sec:fundamentals:classification:pclassification}
Distinct from standard classifiers, probabilistic classifiers focus on estimating the (conditional) class probabilities $\pygivenx(\given{y}{\vec{x}})$ for each class $y$ in $\cY$, for a given input instance $\vec{x} \in \cX$. 
We denote predictions of that kind by $\hpygivenx(\given{y}{\vec{x}})$. As before, training data comes in the form $\cD = \set{(\vec{x}_i, y_i)}_{i=1}^N \subset \cX \times \cY$, where the $(\vec{x}_i, y_i)$ are i.i.d. according to $\pjoint(\cdot)$, and the goal to induce a hypothesis $h_p \from \cX \fromto \setp(\cY)$ with low generalization error (risk):
\begin{equation*}
R_{p}(h_p) = \IE[\loss_p(y, h_p(\vec{x}))] = \inte_{\cX \times \cY} \loss_p(y, h_p(\vec{x})) \, d \, \pjoint(\vec{x}, y) \, .
%\label{eq:prisk}
\end{equation*}
Now, however, instead of comparing a predicted class $\hat{y}$ with a true class $y$, the loss $\loss_p$ compares a predicted probability distribution with $y$\,---\,thus, the loss is a mapping $\loss_p \from \, \cY \times \setp(\cY) \fromto \overline{\IR}$, where $\setp(\cY)$ denotes the set of \acp{PMF} on $\cY$.
The most commonly used loss function $\loss_p(\cdot)$ is the \ac{CCE},  defined as $\loss_{\textscsix{CCE}}(y, \vec{\hat{p}}) \coloneq -\ln( \hat{p}_y)$ \cite[chap.~4]{bishop2006pattern}.
The \ac{CCE} loss (logarithmic proper scoring rule) is widely recognized for its information-theoretic interpretations and practical effectiveness \citep{gneiting2007strictly}.

Since $\cY$ is finite and consists of $\nc$ classes, $\setp(\cY)$ can be represented by the $(\nc-1)$-simplex, i.e., predictions can be represented as probability vectors $h_p(\vec{x}) = \vec{\hat{p}} = (\hat{p}_1, \ldots, \hat{p}_{\nc})$, where $\hat{p}_{\ncs} = \vec{\hat{p}}[\ncs] = \hpygivenx(\given{\ncs}{\vec{x}})$ is the probability assigned to class $\ncs$.
Typically, learning predictors of that kind involves minimizing the \emph{empirical risk}
\begin{equation}
R_{\givens{\text{emp}}{p}}(h_p) = \dfrac{1}{N}\sum_{i=1}^N \loss_p(y_i , h_p(\vec{x}_i)) \, .
\label{eq:perisk}
\end{equation}
In the subsequent discussions, we denote the empirical risk minimizer by $g_p = \argmin_{h_p \in \cH_p} R_{\givens{\text{emp}}{p}}(h_p)$ \cite[chap.~4]{vapnik1991principles, bishop2006pattern}.
In the case where $Y$ is independent of $X$, the \bpmc or \mc (denoted by $g_{p}^{\textnormal{mc}}$) is again the best constant probability predictor, i.e., the one with the lowest risk \eqref{eq:perisk} among all constant predictors.
Obviously, a probabilistic prediction can be used deterministically by selecting the class with the highest predicted probability: $\hat{y} = \argmax_{y \in \cY} \hpygivenx(\given{y}{\vec{x}})$ to induce a deterministic classifier $g$.
This approach minimizes the standard $\zo$ loss on average, emphasizing the importance of accurately identifying the most probable class for optimal performance \cite[chap.~4]{bishop2006pattern}.

\subsubsection{Bayes-optimal Predictor}
\label{sec:fundamentals:bayespredictor}
In \slt, the \emph{\bp} is the optimal classification function $g^{\textnormal{b}} \from \cX \fromto \cY$, which minimizes expected risk \eqref{eq:risk} for a given loss function $\loss(\cdot) \from \cY \times \cY \fromto \IR$:
\begin{align*}
g^{\textnormal{b}}(\vec{x}) &= \argmin_{\hy \in \cY}\sum_{y \in \cY} \loss(y,\hy) \, \product \pygivenx(\given{y}{\vec{x}}) = \argmin_{\hy \in \cY} \IE_{y}[\given{\loss(y,\hat{y})}{\vec{x}}] \, ,
%\label{eq:bayesclassifier}
\end{align*}
where $\IE_{y}[\loss]$ is the expected loss of prediction $\hat{y}$ for $y \in \cY$, and $\pygivenx(\given{y}{\vec{x}})$ is the (conditional) class $y$ probability for input $\vec{x}$ \citep{devroye2013probabilistic}.
For $\zo$ loss, it simplifies to:
\begin{equation}
g^{\textnormal{bc}}(\vec{x}) = \argmax_{y \in \cY} \pygivenx(\given{y}{\vec{x}}) \, .
\label{eq:bayesclassifiercondition}
\end{equation}
It produces the minimum expected loss, known as \emph{\berr}, denoted by $\metricbp{Err}{g^{\textnormal{bc}}}$.
When $\cX$ and $\cY$ are independent of each other, i.e., $\pygivenx(\given{y}{\vec{x}})= \pymarg(y), \forall \vec{x} \in \cX$, it reduces to the \bpmc:
\begin{equation}
g^{\textnormal{bc}}(\vec{x}) \equiv g^{\textnormal{mc}}(\vec{x}) \equiv \argmax_{y \in \cY} \pymarg(y)\, .
\label{eq:marginalclassifier}
\end{equation}
It assigns to each input $\vec{x}$ a class label from the set of labels with the highest marginal probability, as input features are completely uninformative.
Strictly speaking, as the maximum in \eqref{eq:marginalclassifier} is not necessarily unique, the \bpmc may pick any label with the highest probability\,---\; in that case, we nevertheless assume that it picks the same label for every $\vec{x}$, so that it is a constant function. 

\subsection{Quantifying and Detecting Information Leakage}
\label{sec:ilproblem:ilqd}
\Ac{IL} occurs when observable information ($\vec{x} \in \cX$, represented by $X$) is correlated with secret information ($y \in \cY$, represented by $Y$), allowing inference of $y$ from $\vec{x}$ \citep{john2002compression, chatzikokolakis2010statistical}.
To quantify \ac{IL}, we introduce \ac{LAS} $\delta(\cdot)$, evaluating the difference in average penalties of \bpmc and \bp using loss functions ($\loss_{(\cdot)}$) or metrics ($\met_{(\cdot)}$):
\begin{equation*}
\delta(\met_{(\cdot)}) = \card{\met_{(\cdot)}(g^{\textnormal{mc}}) - \met_{(\cdot)}(g^{\textnormal{bc}})}, \, \,
\delta(\loss_{(\cdot)}) = \loss_{(\cdot)}(g^{\textnormal{mc}}) - \loss_{(\cdot)}(g^{\textnormal{bc}}) \, .
\label{eq:las}
\end{equation*}
where $\card{\cdot}$ is used to avoid negative values for accuracy measures.
For \llmis loss function, \ac{LAS} is equal to \textbf{\ac{MI}}, i.e., $\delta(\loss_{ll}) = \mi(X; Y) = \IE[\loss_{ll}(g^{\textnormal{mc}})] - \IE[\loss_{ll}(g^{\textnormal{bc}})]$ and \berr ($\metricbp{Err}{g^{\textnormal{bc}}}$) bounds the \ac{MI}, which serves as the foundation for our \llmi and \midpoint estimation approaches, respectively, as discussed in \Cref{sec:miestimation:approaches}.

In practice, since $\pjoint(.)$ is seldom observed, we approximate \ac{LAS} using empirical risk minimizers ($g_p$ or $g$) as proxies for \bp and $g_{p}^{\textnormal{mc}} $ for \bpmc:
\begin{equation*}
\delta(\met_{(\cdot)}) \approxeq \card{\met_{(\cdot)}(g_{p}^{\textnormal{mc}} ) - \met_{(\cdot)}(g)} , \, \, \delta(\loss_{(\cdot)}) \approxeq \loss_{(\cdot)}(g_{p}^{\textnormal{mc}} ) - \loss_{(\cdot)}(g_p)\, .
\end{equation*}
For losses evaluated using (conditional) class probabilities, $g_p$ minimizing \eqref{eq:perisk} is used as a proxy for the \bp.
However, for decision-based classification losses, $g$ minimizing \eqref{eq:erisk} or the decision rule of $g_p$ is used, with $g$ being a better proxy for \errn or accuracy measure.

\paragraph{\ac{IL} Detection}
\ac{IL} occurs if \ac{LAS} is significantly greater than $0$, i.e., $\delta(\met_{(\cdot)}) \gg 0$ or $\delta(\loss_{(\cdot)}) \gg 0$. 
Thus, \ac{ILD} involves analyzing the learnability of empirical risk minimizers ($g_p$ or $g$) on the dataset $\cD$ used for the system.
Our prior work \citep{ild2022pritha} introduced classification-based \ac{ILD} approaches using an average \errn ($\zo$ loss) to quantify and detect \ac{IL} by analyzing $\delta(\metric{Err})$ and $\delta(\metric{CM})$ using \ac{PTT} and \ac{FET}, respectively.
We also propose \ac{MI}-based \ac{ILD} approaches using the \ac{OTT} on \ac{MI} estimates, with \calmi effectively detecting \ac{IL} in OpenSSL TLS servers (c.f. \Cref{sec:ild:results}).

\section{Mutual Information Estimation}
\label{sec:miestimation}
This section introduces two methods for estimating \ac{MI} in classification datasets.
We compare their performance on synthetic datasets against state-of-the-art approaches including \ac{MINE} \citep{belghazi2018mine}, \ac{GMM} \citep{mariapolo2022effective}, and \pcsoftmax \citep{zhenyue2019rethinking}.
While \ac{MINE} and \ac{GMM} do not explicitly handle imbalance—\ac{MINE} which was initially proposed to estimate \ac{MI} for two real-valued vectors, the classes are binary encoded to estimate \ac{MI} and is already proposed to be assess \ac{IL}, and \ac{GMM} fits per-class Gaussians, which are widely used state-of-the-art approaches \citep{cristiani2020mineild}.
In contrast, \pcsoftmax addresses class imbalance by incorporating label frequency to modify \ac{CCE} loss function to propose \ac{PC-softmax} to estimate \ac{MI} considering the $\pymarg(Y)$, making it a fair and relevant baseline for comparison with \llmi \citep{zhenyue2019rethinking}.
We describe the experimental setup, including the configuration of datasets generated by simulated systems using the \ac{MVN} distribution and evaluation metrics, and summarize the results in \Cref{sec:miestimation:results}.
Our results demonstrate the superior performance of our proposed methods for estimating \ac{MI}.

\subsection{Our Approaches}
\label{sec:miestimation:approaches}
We introduce two methods for estimating \ac{MI} in classification datasets called \llmi, which explicitly targets imbalanced datasets, and \midpoint, which estimates \ac{MI} using only model accuracy, rather than relying solely on standard statistical methods.

\subsubsection{Mid-point Estimation}
\label{sec:miestimation:midpointmi}
The \midpoint approach estimates \ac{MI} by leveraging the relationship between \berr $\metricbp{Err}{g^{\textnormal{bc}}}$ and the conditional entropy $\ent(\given{Y}{X})$ for a classification task \citep{tebbe1968uncertainty}.
The \ac{MI} is estimated using the empirical risk minimizer $g$ of \eqref{eq:erisk} to approximate the \berr. 

\noindent
The conditional entropy $\ent(\given{Y}{X})$ is bounded as
\begin{align*}
	&H_l(\metricbp{Err}{g^{\textnormal{bc}}}, \nc) \leq \ent(\given{Y}{X}) \leq H_u(\metricbp{Err}{g^{\textnormal{bc}}}, \nc) \, \\
	&H_l(\metricbp{Err}{g^{\textnormal{bc}}}, \nc) = \log(\ncs) + \ncs (\ncs + 1) \left(\log\left(\frac{\ncs+1}{\ncs}\right)\right) \left(\metricbp{Err}{g^{\textnormal{bc}}} - \frac{\ncs-1}{\ncs}\right) \\
	&H_u(\metricbp{Err}{g^{\textnormal{bc}}}, \nc) = H_2(\metricbp{Err}{g^{\textnormal{bc}}}) + \metricbp{Err}{g^{\textnormal{bc}}} \cdot \log(\nc-1) \, ,
	%\label{eq:entlowerbound}
\end{align*}
where $H_l(\cdot)$ derived by \citet{hellman1970probability} is valid for $\frac{1}{\ncs+1} \leq 1- \metricbp{Err}{g^{\textnormal{bc}}} \leq \frac{1}{\ncs}\, ,\ncs = 1, \ldots, \nc-1$ and $H_u(\cdot)$ derived by \citet{fano1961transmission} uses the binary cross-entropy function, $H_2(a) = -a\log(a) - (1-a) \log(1-a)$.

\noindent
Plugging in $H_l(\cdot)$ and $H_u(\cdot)$ in \eqref{eq:mi_y}, the bounds on \ac{MI} the lower bound $F_l(\cdot)$ and the upper bound $F_u(\cdot)$ with respect to \berr and the number of classes $\nc$ is derived as
\begin{align*}
&\ent(Y) - H_u(\metricbp{Err}{g^{\textnormal{bc}}}, \nc) \leq \mi(X; Y) \leq \ent(Y) - H_l(\metricbp{Err}{g^{\textnormal{bc}}}, \nc) \\
&F_l(\metricbp{Err}{g^{\textnormal{bc}}}, \nc) \leq \mi(X; Y) \leq F_u(\metricbp{Err}, {g^{\textnormal{bc}}}, \nc) .
\label{eq:mibound}
\end{align*}
\ac{MI} is estimated as
\begin{equation}
\hmi(X; Y) \approxeq \frac{F_u(\metricbp{Err}{g}, \nc) + F_l(\metricbp{Err}{g}, \nc)}{2} \, ,
\label{eq:midpoint}
\end{equation}
where $g$ is the \emph{empirical risk minimizer} of \eqref{eq:erisk} and serves as a proxy of the \bp, as $\metricbp{Err}{g^{\textnormal{bc}}} \approxeq \metricbp{Err}{g}$.

\begin{figure}[!htb]
	\centering
	\includegraphics[width=\textwidth]{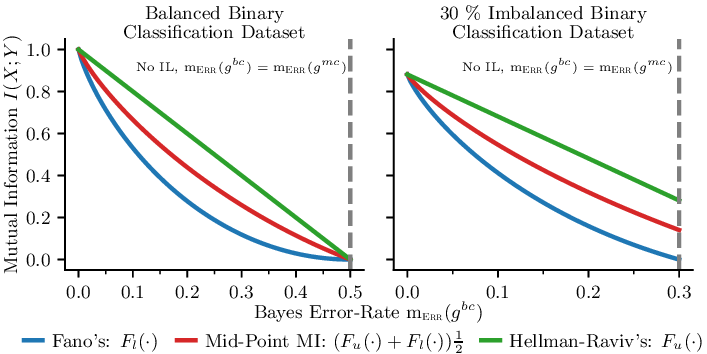}
        \caption{Bounding \ac{MI} (y-axis) with respect to \berr (x-axis) for a balanced and imbalanced binary classification dataset with \SI{30}{\%} of minority classes (\emph{class imbalance} level $r=0.3$).
        The upper bound $F_u(\cdot)$ is derived using $H_l(\cdot)$ by \citet{hellman1970probability} and the lower bound $F_l(\cdot)$ is defined using the upper bound $H_u(\cdot)$ derived by \citet{fano1961transmission}.
        The gray lines represent the case of no \ac{IL}, where the Bayes-optimal predictor converges to the marginal Bayes predictor.}
    \label{fig:midpoint}
\end{figure}

\paragraph{Overestimation in Imbalanced Data}
To illustrate the issue of \ac{MI} overestimation by the \midpoint approach—leading to false positives in \ac{IL} detection—we visualize \ac{MI} estimated using \midpoint in a \emph{non-vulnerable} system that generates imbalanced binary classification data with a class  \emph{imbalance level} of \SI{30}{\%} ($r = 0.3$, defined in \ref{par:syntheticsystems:method:pdfs:imbalance}). 
The \emph{non-vulnerable} system corresponds to the condition where the Bayes-optimal predictor converges to the marginal Bayes predictor, as discussed in \Cref{sec:ilproblem:ilqd}.
While \citet{ming2013fano} relates metrics like \ac{BER} to the conditional entropy $\ent(\given{Y}{X})$ in binary classification, such relationships do not extend naturally to the multi-class case.

\subsubsection{\llmi Estimation}
\label{sec:miestimation:llmi}
We propose the \llmi approach for MI estimation to address the overestimation and false positives of the \midpoint approach and the false negatives of baseline \ac{ILD} approaches.
This approach uses the empirical risk minimizer $g_p$ minimizing \eqref{eq:perisk} to approximate the \llmis of the \bp.

The \ac{MI} between $X$ and $Y$ in \eqref{eq:mi_y} is defined as 
\begin{align}
	\mi(X; Y) &=-\ent(\given{Y}{X}) + \ent(Y) \nonumber \\ 
	&=\inte_{\vec{x} \in \cX} \pxmarg(\vec{x}) \sum_{y \in \cY} \pygivenx(\given{y}{\vec{x}}) \log \left(\pygivenx(\given{y}{\vec{x}})\right) d\vec{x} -\sum_{y \in \cY} \pymarg(y) \log {\left(\pymarg(y)\right)} \nonumber \\ 
	%&=-\text{Expected \llmis of \bp} + \text{Expected \llmis of \bpmc} \\
	%&= -\IE[\loss_{ll}(\text{\bp})] + \IE[\loss_{ll}(\text{\bpmc})] \nonumber \\ 
	&= -\IE[\loss_{ll}(g^{\textnormal{bc}})] + \IE[\loss_{ll}(g^{\textnormal{mc}})] \approxeq - \IE[\loss_{ll}(g_p)] + \IE[\loss_{ll}(g_{p}^{\textnormal{mc}} )] \, , \nonumber 
\end{align}
where $\IE[\loss_{ll}(\cdot)]$ represents the expected \llmis.
The conditional entropy $\ent(\given{Y}{X})$ equals the expected \llmis of the \bp $g^{\textnormal{bc}}$, while the entropy of $Y$ corresponds to the expected \llmis of the \bpmc $g^{\textnormal{mc}}$, reaffirming \ac{MI} as a special case of \ac{LAS} (c.f. \Cref{sec:ilproblem:ilqd}).

\paragraph{Log-loss}
\label{sec:miestimation:llmi:ll}
For probabilistic classifiers $g_p$, \ac{CCE} is used to obtain estimated conditional class probabilities, forming a vector $\vec{\hat{p}} = (\hat{p}_1, \ldots, \hat{p}_{\nc})$ for each input $\vec{x}$. 
The \llmis for $g_p$ is defined in \citet[chap.~4]{bishop2006pattern} as
\begin{equation}
	\loss_{ll}(y, \vec{\hat{p}}) = -\sum_{\ncs = 1}^{\nc} \hat{p}_{\ncs} \log(\hat{p}_{\ncs}) = -\sum_{\ncs = 1}^{\nc} g_{p}(\vec{x})[\ncs] \log(g_{p}(\vec{x})[\ncs]) \, .\nonumber
\end{equation}
It reaches its minimum value of $0$ when one class dominates, i.e., $\hat{p}_{\ncs} \approxeq 1$ and $\hat{p}_j \approxeq 0$ for all $j \in [\nc] \setminus \set{\ncs}$, and its maximum $\log(\nc)$ when class probabilities are uniform, i.e., $\hat{p}_{\ncs} = \sfrac{1}{\nc}$ for all $\ncs \in [\nc]$, implying that high classification uncertainty increases the \llmis, making it suitable for estimating the conditional entropy $\ent(\given{Y}{X})$.

\paragraph{Marginal \bp}
\label{sec:miestimation:llmi:mc}
It is estimated from the class distribution in the dataset $\cD = \set{(\vec{x}_i, y_i)}_{i=1}^N$ using \eqref{eq:perisk} and is denoted by $g_{p}^{\textnormal{mc}} $.
The class probabilities for each input $\vec{x}$ are $g_{p}^{\textnormal{mc}} (\vec{x}) = \vec{\hat{p}}_{mc} = (\hat{p}_1, \ldots, \hat{p}_{\nc})$, where $\hat{p}_{\ncs} = \tfrac{\card{\set{(\vec{x}_i, y_i) \in \cD \; \lvert \; y_i=\ncs}}}{\card{\cD}}$ is the fraction of instances for class $\ncs$. 
The \llmis of the \bpmc suitably estimates the entropy of $Y$, i.e., $\ent(Y) \approxeq \loss_{ll}(y, \vec{\hat{p}}_{mc})$, ranging from $0$ to $\log(\nc)$, with the maximum acquired for a balanced dataset.

\paragraph{\ac{MI} estimation}
\label{sec:miestimation:llmi:estimation}
We approximate the expected \llmis of the \bp and \bpmc by evaluating the \llmis of $g_p$ and $g_{p}^{\textnormal{mc}} $ for the dataset $\cD = \set{(\vec{x}_i, y_i)}_{i=1}^N$ as
\begin{align*}
	&\IE[\loss_{ll}(g_p)] = \dfrac{1}{N}\sum_{i=1}^N \loss_{ll}(y_i, g_p(\vec{x}_i)) = -\dfrac{1}{N}\sum_{i=1}^N\sum_{\ncs = 1}^{\nc} g_{p}(\vec{x}_i)[\ncs] \log(g_{p}(\vec{x}_i)[\ncs]) \\
	&\IE[\loss_{ll}(g_{p}^{\textnormal{mc}} )] = \dfrac{1}{N}\sum_{i=1}^N \loss_{ll}(y_i, g_{p}^{\textnormal{mc}} (\vec{x}_i)) = -\sum_{\ncs = 1}^{\nc} \vec{\hat{p}}_{mc}[\ncs] \log(\vec{\hat{p}}_{mc}[\ncs]).
\end{align*}

The \ac{MI} is then estimated as
\begin{equation}
	\hmi(X; Y) \approxeq \sum_{\ncs = 1}^{\nc} \left(\dfrac{1}{N}\sum_{i=1}^N g_{p}(\vec{x}_i)[\ncs] \log(g_{p}(\vec{x}_i)[\ncs]) - \vec{\hat{p}}_{mc}[\ncs] \log(\vec{\hat{p}}_{mc}[\ncs]) \right).
	\label{eq:llmiest}
\end{equation}

\paragraph{Classifier Calibration}
Despite the sound theoretical grounding of the proper scoring rule loss function, \ac{CCE} often yields poorly calibrated probabilities in practice, with neural networks tending to be overconfident, especially on imbalanced datasets \citep{szegedyls}.
Post-hoc \emph{calibration} techniques address this by learning mappings from predicted to improved probabilities using validation data \citep{silva2023calibration}.
To enhance \llmi estimation accuracy, we employ calibration techniques like Isotonic Regression (\ircalmi), Platt's Scaling (\pscalmi), Beta Calibration (\betacalmi), Temperature Scaling (\tscalmi), and Histogram Binning (\hbcalmi), referred collectively as \calmi. \citep{silva2023calibration}.
These calibration techniques are pivotal for enhancing \llmi, and enabling accurate \ac{MI} estimation—thereby avoiding the false positives often observed with the \midpoint approach.

\subsection{Empirical Evaluation}
\label{sec:miestimation:experiments}
\begin{figure}[htbp]
	\centering
	\includegraphics[width=\textwidth]{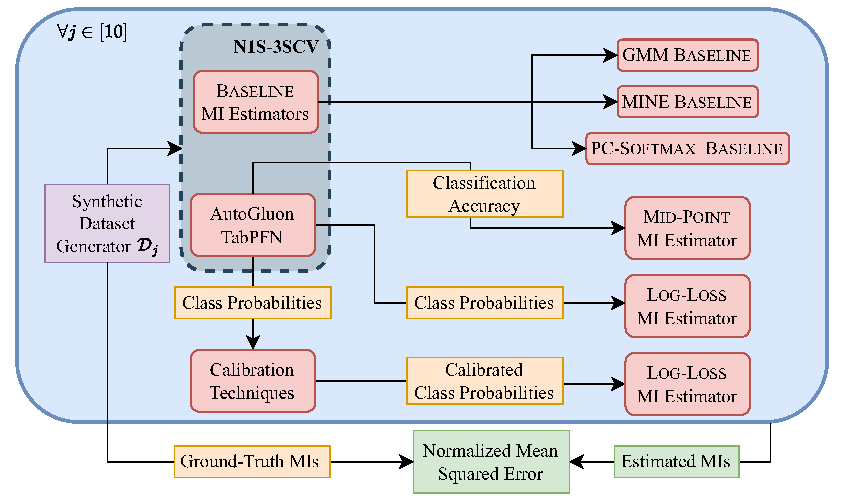}
	\caption{Experimental setup for evaluating \ac{MI} estimation methods}
	\label{fig:mitechniques}
\end{figure}

This section outlines the evaluation process for our \ac{MI} estimation methods compared to baselines, as illustrated in \Cref{fig:mitechniques}.
Our goal is to assess the generalization capabilities of these approaches under various conditions, including the number of classes ($\nc$), input dimensions ($d$), class imbalance ($r$), and noise level ($\epsilon$), using datasets generated by systems with different configurations ($\gentype, \nc, d, r, \epsilon$) outlined in \Cref{tab:syntheticdatasetoverview}.
Overall and generalization performance results are discussed in \Cref{sec:miestimation:results}.

\begin{table}[!htb]
	\centering
	\caption{Overview of the synthetic datasets for \ac{MI} estimation experiments}
	\label{tab:syntheticdatasetoverview}
	\sisetup{
		table-figures-decimal = 1,
		table-figures-exponent = 1,
		table-number-alignment = center
	}
	\begin{adjustbox}{width=\textwidth,center}
		\begin{tabular}{c|c|c|c|c|c}
			\toprule
			\multicolumn{6}{c}{\Ac{MVN} Perturbation \& \Ac{MVN} Proximity Synthetic Datasets} \\
			\midrule
                \makecell{Dataset\\Type} & \makecell{Generation\\Method ($\gentype$)} & \makecell{Input\\Dimensions ($d$)} & {Classes ($\nc$)} & \makecell{Noise Level/\\Flip Percentage ($\noise$)} & \makecell{Class Imbalance ($r$)} \\
			\midrule
			Balanced & NA & {\set{2, 4, \ldots, 20}} & {\set{2, 4, \ldots, 10}} & {\set{0.0, 0.1, \ldots, 1.0}} & NA \\
			Imbalanced Binary-class & {\text{Minority}} & {5} & {2} & {\set{0.0, 0.1, \ldots, 1.0}} & {\set{0.05, 0.1, \ldots, 0.5}} \\
			Imbalanced Multi-class & {\text{Minority}, \text{Majority}} & {5} & {5} & {\set{0.0, 0.1, \ldots, 1.0}} & {\set{0.02, 0.04, \ldots, 0.2}} \\
			\bottomrule
		\end{tabular}
	\end{adjustbox}
\end{table}

\subsubsection{Synthetic Datasets}
\label{sec:miestimation:experiments:datasets}
To evaluate our \ac{MI} estimation methods, we generated synthetic datasets by simulating real-world systems using the \ac{MVN} distribution, allowing for straightforward calculation of ground truth \ac{MI}, discussed in \ref{asec:syntheticsystems:method}.
\ac{MVN} perturbation and proximity techniques were used to introduce noise and simulate systems with varying levels of vulnerability, as detailed in \ref{asec:syntheticsystems:noise}.
The perturbation technique introduces noise by flipping a percentage of class labels, and the proximity technique reduces the distance between the mean vectors of the \ac{MVN} for each class, causing the generated Gaussians to overlap.
To simulate the realistic scenarios, we define the \bemph{vulnerable systems} as those with noise levels $\epsilon \in [0.0, 0.90)$, where side-channel signals remain distinguishable, and \bemph{non-vulnerable systems} as those with $\epsilon \in [0.90, 1.0]$, where high noise renders secret information statistically indistinguishable using the side-channel observable information.

\begin{figure}[htb]
	\centering
	\includegraphics[width=\textwidth]{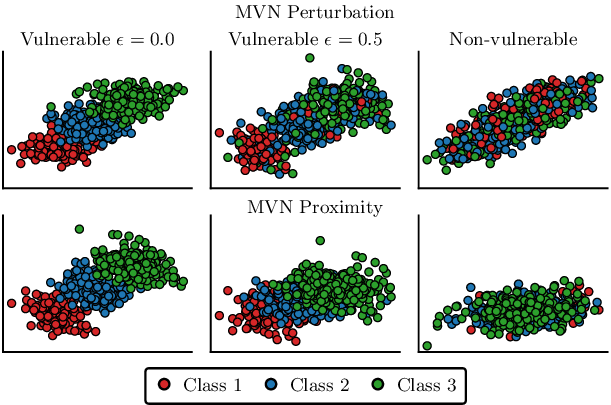}
            \caption{Visualization of three-class synthetic datasets in $2$-D space, generated using \ac{MVN} perturbation and proximity-based techniques to simulate \bemph{vulnerable} systems ($\epsilon = 0$, $0.5$) and a \bemph{non-vulnerable} system ($\epsilon = 1.0$).}
	\label{fig:synthetic_datasets}
\end{figure}
This assumption is supported by the \textbf{Marvin \ac{SCA}} \citep{kario2024marvin}, which shows that even noisy side-channels can be exploited using statistical tests and system-level optimizations.
\Cref{fig:synthetic_datasets} illustrates two-dimensional data points with class \emph{imbalance level} ($r = 0.3$), generated using both \ac{MVN} perturbation and proximity-based techniques. 
The datasets include \bemph{vulnerable} systems at noise levels $\epsilon = 0$ and $\epsilon = 0.5$, and a \bemph{non-vulnerable} system at $\epsilon = 1.0$.

\begin{figure}[htb]
	\centering
	\includegraphics[width=\textwidth]{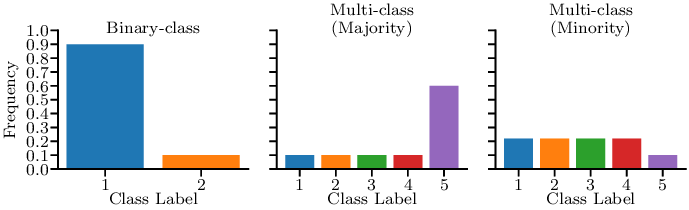}
	\caption{Classification dataset class distribution with \SI{10}{\percent} of \emph{class imbalance} level ($r=0.1$)}
	\label{fig:imbalanced_datasets}
\end{figure}
We introduce \emph{class imbalance} ($r$) using \emph{Minority} and \emph{Majority} sampling strategies to define the output prior $\pymarg(\cdot)$, as described in \ref{par:syntheticsystems:method:pdfs:imbalance}, with example class distributions for each case of imbalanced binary-class and multi-class datasets illustrated in \Cref{fig:imbalanced_datasets}.
To evaluate \ac{MI} estimation accuracy of different approaches under varying class distributions (class imbalance levels $r$), we generated datasets for balanced, imbalanced binary-class, and multi-class scenarios.
Each dataset configuration varies in input dimension ($d$), number of classes ($\nc$), class imbalance level ($r$), and noise level ($\epsilon$), as summarized in \Cref{tab:syntheticdatasetoverview}.
All datasets were independently generated across the full noise range ($\epsilon \in [0.0, 1.0]$) to simulate both \bemph{vulnerable} ($\epsilon \in [0.0, 0.90)$) and \bemph{non-vulnerable} ($\epsilon \in [0.90, 1.0]$) systems.

\subsubsection{Experimental Setup}
\label{sec:miestimation:experiments:setup}
We evaluate each \ac{MI} estimation approach on synthetic datasets generated for configurations outlined in \Cref{tab:syntheticdatasetoverview}, as shown in \Cref{fig:mitechniques}.
For each configuration ($\gentype, \nc, d, r, \epsilon$), we generate $10$ datasets ($\cD_j$) using different seeds $j \in [10]$ and evaluate the performance using \ac{NMAE}.
Using nested cross-validation with \ac{HPO}, we split $\cD_j$ into \SI{70}{\percent} training and \SI{30}{\percent} test datasets.
\ac{HPO} involves $100$ function evaluations using \ac{MCCV} with $3$ splits on the training set, reserving \SI{30}{\percent} for validation, denoted as \enquote{N$1$S-$3$SCV}, as shown in \Cref{fig:mitechniques}.
Objective functions for \ac{HPO} include \ac{BER} for \pcsoftmax, \autogluon, and \tabpfn, \ac{AIC} for \ac{GMM}, and \ac{MSE} for \ac{MINE} with parameter ranges provided in \Cref{tab:ranges}.
We identify the best-performing pipeline from running \autogluon for $1800$ seconds and model using \ac{HPO} on \ac{GMM}, \ac{MINE},  \pcsoftmax, and \tabpfn using validation loss or accuracy.
The estimated \ac{MI} of this model on the dataset $\cD_j$, is compared with respect to the ground truth \ac{MI} in \eqref{eq:gtmvn_mi}, denoted by $\hat{I}_j$ and $I_j$, using the \ac{NMAE} defined in \eqref{eq:nmae}.

\subsubsection{Evaluation Metric}
\label{sec:miestimation:experiments:metric}
We assess generalization using normalized \ac{MAE} (\ac{NMAE}), computed by dividing \ac{MAE} by the entropy of $Y$, i.e., $\ent_{Y}(\gentype, \nc, r) = -\sum_{\ncs=1}^{\nc} \pymarg(\ncs) \log \left(\pymarg(\ncs)\right)$, since \ac{MI} ranges from $0$ to $\log(\nc)$.
The \ac{NMAE} metric for $10$ ground-truth and estimated \ac{MI} values is:
\begin{equation}
\metric{NMAE}(\vec{I}, \vec{\hat{I}}) = \frac{1}{10}\sum_{j=1}^{10} \frac{\card{\vec{I}[j] - \vec{\hat{I}}[j]}}{\ent_{Y}(\gentype, \nc, r)} =\sum_{j=1}^{10} \frac{\card{{I}_j - \hat{I}_j}}{10 \product \ent_{Y}(\gentype, \nc, r)} \, .
\label{eq:nmae}
\end{equation}

\subsection{Results}
\label{sec:miestimation:results}
This section discusses the overall performance of various \ac{MI} estimation methods on systems simulated using perturbation and proximity techniques. 
We also analyze the generalization capabilities of selected approaches across various factors, including the number of classes ($\nc$), input dimensions ($d$), class imbalance ($r$), and noise levels ($\epsilon$).
Our analysis covers balanced, binary-class, and imbalanced multi-class datasets, as detailed in \Cref{tab:syntheticdatasetoverview}.
We examine datasets generated by simulated \bemph{vulnerable systems} with noise levels of \SI{0}{\percent} ($\epsilon = 0.0$, representing a fully compromised system) and \SI{50}{\percent} ($\epsilon = 0.5$, representing a compromised system with ineffective countermeasures against \acp{SCA}).
In contrast, datasets from \bemph{non-vulnerable systems} are simulated with noise levels $\epsilon \in [0.90, 1.0]$, representing systems robust even against advanced attacks such as the Marvin \ac{SCA}, which can exploit noisy information to reveal secrets \citep{kario2024marvin}.
This setup highlights the importance of accurate \ac{MI} estimation for effective \ac{ILD}.
Additionally, to assess accurate \ac{MI} estimation for \ac{ILD} in imbalanced datasets from realistic systems, we examine the binary and multi-class imbalanced datasets generated by various vulnerable systems ($\epsilon \in [0.00, 0.90)$) and two non-vulnerable systems with $\epsilon \in \set{0.90, 1.0}$.

\subsubsection{Overall Performance}
\label{sec:miestimation:results:overall}
\begin{figure}[!htb]
	\begin{subfigure}{\textwidth}
		\centering
		\includegraphics{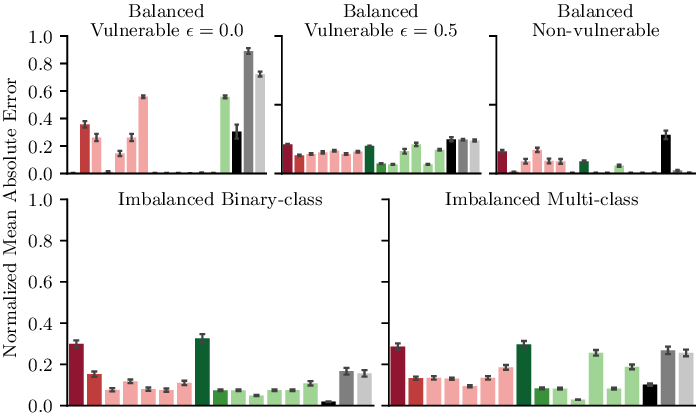}
		\caption{Performance on systems simulated using \ac{MVN} perturbation technique}
		\label{fig:mvnperturbation_overall}
	\end{subfigure}
	\begin{subfigure}{\textwidth}
		\centering
		\includegraphics{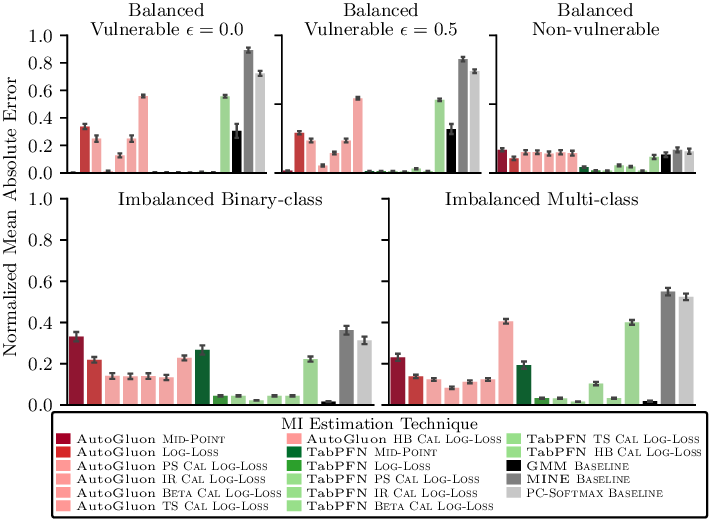}
		\caption{Performance on systems simulated using \ac{MVN} proximity technique}
		\label{fig:mvnproximity_overall}
	\end{subfigure}
	\caption{Overall \ac{NMAE} of \ac{MI} estimation methods on synthetic datasets}
	\label{fig:overall_results}
\end{figure}

We present the performance of various \ac{MI} estimation methods using bar charts showing mean and \ac{SE} of \ac{NMAE} on systems simulated by \ac{MVN} perturbation and proximity techniques, as shown in \Cref{fig:mvnperturbation_overall,fig:mvnproximity_overall}, respectively.

\paragraph{\ac{MVN} Perturbation}
\label{sec:miestimation:results:overall:mvnperturbation}
The results depicted in \Cref{fig:mvnperturbation_overall} indicate that \tabpfn \ircalmi consistently excels in estimating \ac{MI}, and methods using \autogluon significantly underperform compared to those using \tabpfn.
For balanced datasets, \tabpfn \ircalmi shows exceptional performance ($\approx 0.003$) and low \ac{SE} of $0.001$.
For more vulnerable systems with noise ($\epsilon = 0.5$), \tabpfn \ircalmi performs less effectively ($\approx 0.212 \pm 0.012$), while \tabpfn \pscalmi and \betacalmi perform better ($\approx 0.067 \pm 0.003$).
The \hbcalmi method does not enhance the \llmi estimation for both \ac{AutoML} tools.
For systems generating \emph{imbalanced binary-class} datasets, the \ac{GMM} shows exceptional performance ($\approx 0.019 \pm 0.002$), with \tabpfn \llmi and \tabpfn \ircalmi also performing well ($\approx 0.05$).
For \emph{imbalanced multi-class} datasets, \tabpfn \ircalmi leads ($\approx 0.029 \pm 0.001$), with the \ac{GMM} also performing well ($\approx 0.10 \pm 0.002$), demonstrating strong adaptability for imbalanced datasets.

\paragraph{\ac{MVN} Proximity}
\label{sec:miestimation:results:overall:mvnproximity}
Overall, \ac{MI} estimation is more straightforward on datasets generated using proximity than on ones using the perturbation technique, as it introduces more complexity (\ref{asec:syntheticsystems}).
However, performance trends are similar to \ac{MVN} perturbation datasets: \tabpfn \ircalmi outperforms, while \autogluon performs worse.
For systems generating \emph{balanced} datasets, \tabpfn \ircalmi performs exceptionally well for vulnerable systems ($\approx 0.003$ for $\epsilon = 0.0$ and $\approx 0.010$ for $\epsilon = 0.5$) with low variance.
For non-vulnerable systems, \tabpfn \pscalmi and \tscalmi perform best ($\approx 0.016$).
For systems generating \emph{imbalanced binary-class} datasets, the \ac{GMM} and \tabpfn \ircalmi are top performers ($\approx 0.017 \pm 0.001$).
In \emph{imbalanced multi-class} datasets, \tabpfn \ircalmi maintains superior performance ($\approx 0.016$), while the \ac{GMM} also performs well ($\approx 0.017$).

\paragraph{Summary}
\label{sec:miestimation:results:overall:summary}
The \tabpfn approaches, particularly \ircalmi, consistently excel in \ac{MI} estimation, with \autogluon approaches significantly underperforming compared to those using \tabpfn.
Calibration techniques (\calmi) generally do not significantly impact \tabpfn \llmi's performance for balanced datasets but enhance \autogluon \llmi's performance in vulnerable datasets, indicating \autogluon's poor calibration and \tabpfn's well-calibrated class probabilities.
However, these techniques can sometimes worsen \autogluon \llmi's performance in non-vulnerable systems, leading to \ac{MI} overestimation and potential false positives in \ac{ILD}, as shown in \ref{asec:generalization:ild:autogluon}.

\subsubsection{Generalization Performance}
\label{sec:miestimation:results:gen}
\begin{figure}[!htb]
	\centering
	\includegraphics[width=\textwidth]{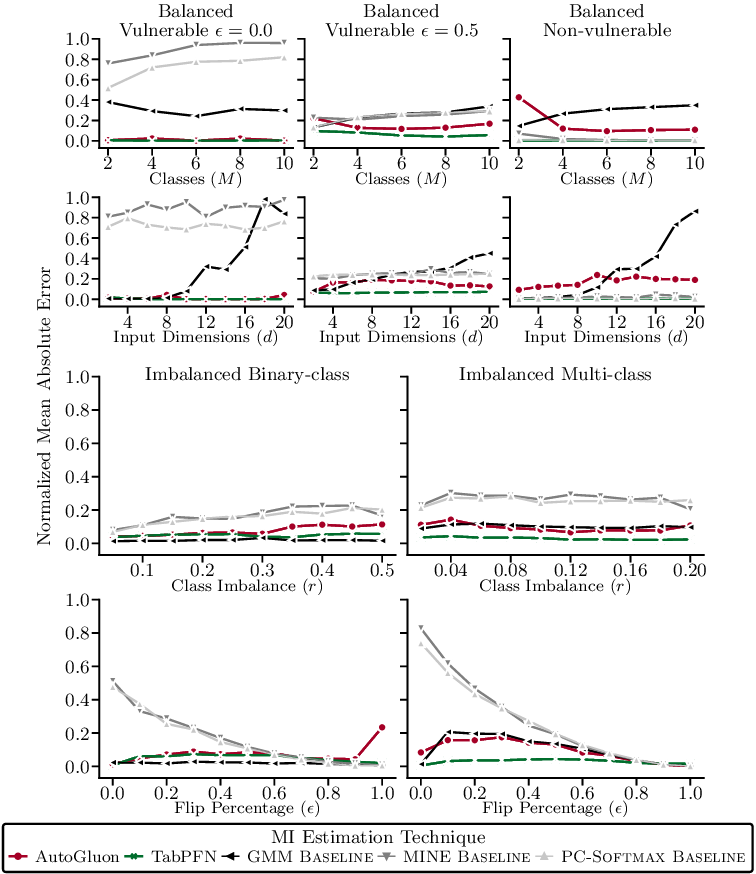}
	\caption{Generalizability of \ac{MI} estimation methods on \ac{MVN} perturbation systems}
	\label{fig:mvnperturbation_best}
\end{figure}

\begin{figure}[!htb]
	\centering
	\includegraphics[width=\textwidth]{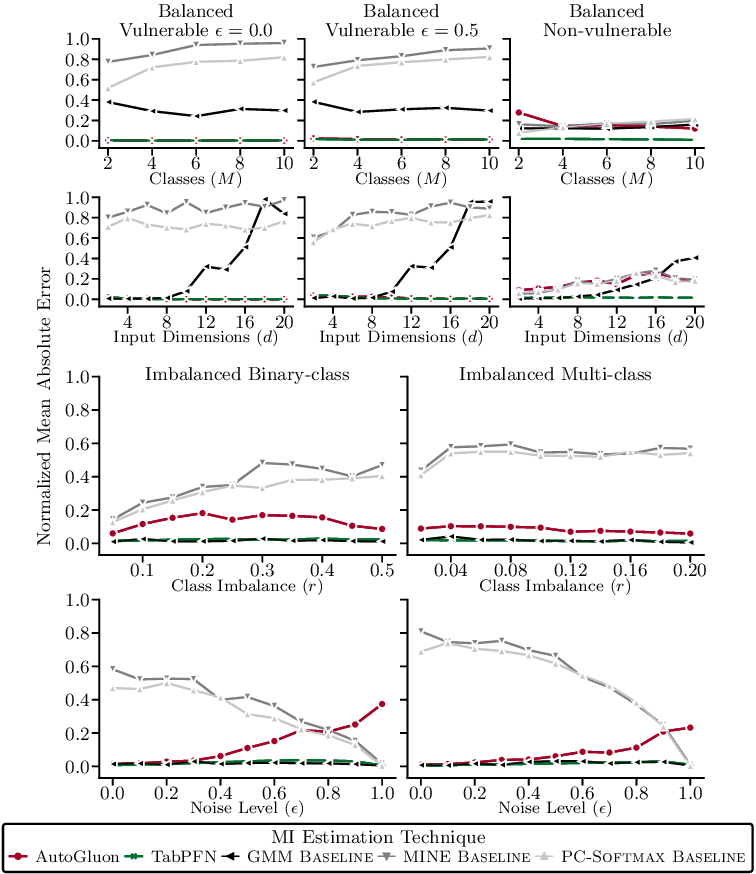}
	\caption{Generalizability of \ac{MI} estimation methods on \ac{MVN} proximity systems}
	\label{fig:mvnproximity_best}
\end{figure}

This section examines the generalization capabilities of different \ac{MI} estimation methods relative to baselines, identifying the best-performing methods using \autogluon and \tabpfn on balanced (non-vulnerable systems with noise levels of $\set{0.90, 1.0}$ and vulnerable systems with noise levels $0.0$ and $0.5$), and imbalanced binary-class and multi-class datasets containing datasets generated by various vulnerable and non-vulnerable systems using \ac{NMAE}, as detailed in \Cref{sec:miestimation:results:overall}.
We assess generalization based on the number of classes ($\nc$) and input dimensions ($d$) by aggregating \ac{NMAE} across dimensions for each class ($\nc \in [2,10]$) and classes for each dimension ($d \in [2,20]$).
For class imbalances, we aggregate \ac{NMAE} across noise levels for each imbalance parameter ($r \in [0.05, 0.5]$ for binary-class and $r \in [0.02, 0.2]$ for multi-class datasets). 
Similarly, we assess generalization for noise levels ($\epsilon \in [0.0, 1.0]$) by aggregating \ac{NMAE} across class imbalances.
\Cref{fig:mvnperturbation_best,fig:mvnproximity_best} visualize these results with line plots, showing \ac{NMAE} on systems simulated using \ac{MVN} perturbation and proximity techniques, respectively.

\paragraph{Number of Classes ($\nc$)}
\label{sec:miestimation:results:gen:classes}
The top row of \Cref{fig:mvnperturbation_best,fig:mvnproximity_best} shows \ac{MI} estimation methods' generalization across datasets with varying numbers of classes ($\nc \in [2,10]$) on the X-axis in systems simulated using \ac{MVN} perturbation and proximity techniques, respectively.
In \emph{vulnerable} and \emph{non-vulnerable} systems, \autogluon and \tabpfn show high accuracy (lower \ac{NMAE}) with increasing classes, with \tabpfn outperforming \autogluon, especially at \SI{50}{\percent} noise ($\epsilon = 0.5$).
Baselines degrade (\ac{NMAE} increases) with the increase in the number of classes.
In \emph{vulnerable} systems, \ac{MINE} is the least accurate, whereas in \emph{non-vulnerable} systems, both \ac{MINE} and \pcsoftmax outperform \ac{GMM}.

\paragraph{Input Dimensions ($d$)}
\label{sec:miestimation:results:gen:features}
The second row of \Cref{fig:mvnperturbation_best,fig:mvnproximity_best} shows \ac{MI} estimation methods' generalization across datasets with varying input dimensions ($d \in [2, 20]$) on the X-axis in systems simulated \ac{MVN} perturbation and \ac{MVN} proximity techniques, respectively.
In \emph{vulnerable} and \emph{non-vulnerable} systems, \autogluon and \tabpfn improve notably with more dimensions, with \tabpfn leading, specifically in systems with \SI{50}{\percent} noise.
Baselines, especially \ac{GMM}, deteriorate significantly with more dimensions, indicating that even deep \acp{MLP} struggle with high-dimensional datasets.
\ac{MINE} and \pcsoftmax perform well in \emph{non-vulnerable} systems simulated using perturbation but deteriorate in ones using the proximity technique.

\paragraph{Class Imbalance ($r$)}
\label{sec:miestimation:results:gen:imbalance}
In the third row of \Cref{fig:mvnperturbation_best,fig:mvnproximity_best} shows \ac{MI} estimation methods' generalization across datasets with varying class imbalances ($r \in [0.05, 0.5]$ for binary-class and $r \in [0.02, 0.2]$ for multi-class) on X-axis, in imbalanced datasets generated using \ac{MVN} perturbation and \ac{MVN} proximity technique, respectively.
In systems generating imbalanced datasets, \tabpfn and \ac{GMM} show high accuracy and stay unaffected, especially in systems generated using perturbation techniques.
\autogluon, \ac{MINE}, and \pcsoftmax are deteriorating, notably beyond $0.25$ \emph{class imbalance} level for systems generating \emph{binary class imbalanced} datasets using perturbation techniques.
While for systems generating \emph{imbalanced multi-class} datasets, \autogluon improves as class imbalance decreases, \ac{MINE} and \pcsoftmax remain relatively constant with a notable drop at $0.04$ \emph{class imbalance} level.

\paragraph{Noise Level ($\epsilon$)}
\label{sec:miestimation:results:gen:noise}
The last row of \Cref{fig:mvnperturbation_best,fig:mvnproximity_best} shows \ac{MI} estimation methods' generalization across datasets with varying noise levels ($\epsilon \in [0.0, 1.0]$) on the X-axis, in imbalanced datasets generated using \ac{MVN} perturbation and \ac{MVN} proximity techniques, respectively.
In systems generating imbalanced datasets, \tabpfn and \ac{GMM} demonstrate high accuracy across all noise levels, particularly in systems generated using the perturbation technique.
\autogluon deteriorates with increasing noise, while \ac{MINE} and \pcsoftmax improve significantly.

\paragraph{Summary}
\label{sec:miestimation:results:gen:summary}
\tabpfn, particularly \ircalmi, consistently demonstrates remarkable generalization in \ac{MI} estimation across various factors for systems simulated using both \ac{MVN} perturbation and proximity techniques.
\tabpfn and \autogluon show strong generalization performance across different \emph{numbers of classes ($\nc$) and input dimensions ($d$)}, with baselines performing worse as they increase.
\ac{GMM} struggles with high-dimensional datasets.
\ac{MINE} shows the lowest generalization capability, suggesting its unsuitability for estimating \ac{MI} in classification datasets.
\tabpfn and \ac{GMM} exhibit robust generalization across various \emph{class imbalances ($r$) and noise levels ($\epsilon$)}.
\ac{GMM} handles noise and imbalance well in low-dimensional datasets ($d=5$) but struggles with high-dimensional ones.
Conversely, \autogluon, \ac{MINE}, and \pcsoftmax show weaker generalization with respect to class imbalances ($r$) and noise levels ($\epsilon$).

\section{Information Leakage Detection}
\label{sec:ild}
This section introduces the \ac{ILD} process with various classification-based and \ac{MI}-based \ac{ILD} approaches.
We describe the experimental setup and \ildds descriptions used to evaluate \acl{ILD} approaches for detecting side-channel leakages  (\acp{IL}) through time delays, countering Bleichenbacher's attacks on OpenSSL TLS servers, and summarize the results with \ref{asec:generalization:ild} providing a detailed analysis.
Our results conclude that our \ac{MI}-based \ac{ILD} approach using the calibrated \llmis (\calmi) outperforms state-of-the-art methods.

\subsection{\ac{ILD} Methodology}
\label{sec:ild:process}
\Cref{fig:ild_combined} illustrates the process of detecting \ac{IL} in systems generating classification datasets using various \ac{ILD} approaches.
To enhance \ac{IL} detection confidence, we apply Holm-Bonferroni correction on $p$-values obtained using statistical tests on performance estimates of the top-$10$ \ac{AutoML} models or pipelines from \autogluon and \tabpfn obtained through \ac{HPO}.

\subsubsection{\ac{ILD} Approaches}
\label{sec:ild:process:approaches}
We divide the \ac{ILD} approaches based on using \ac{MI} and classification metrics to detect \ac{IL} in a system into \ac{MI}-based and classification-based methods.
We propose using the \ac{AutoML} tools \autogluon and \tabpfn, as described in \ref{asec:automl}, to accurately estimate empirical risk minimizers ($g$ or $g_p$), which can serve as an appropriate proxy for \bp.
We propose applying statistical tests to the \ac{LAS} calculated using \llmi, accuracy, and \ac{CM} of the top-performing \ac{AutoML} models or pipelines ($g$ or $g_p$) to obtain a $p$-value.
All statistical tests in our setup follow a common framework, where the null hypothesis $H_0(\cdot)$ assumes no \ac{IL}, and the alternative $H_1(\cdot)$ indicates its presence. 
A $p$-value below the significance level $\alpha$ leads to rejecting $H_0$, thereby inferring the presence of \ac{IL}.
The threshold \textbf{$\alpha = 0.01$} is particularly appropriate for avoiding false positives using the Holm-Bonferroni correction when testing multiple hypotheses to ensure high-confidence detection in security-sensitive settings \citep{ild2022pritha,holm1979simple}. 
We propose to use $\alpha = 0.01$ for the Holm-Bonferroni correction, tuned with the cutoff parameter $\tau = 5$ from our prior work \citep{ild2022pritha}.

\paragraph{\ac{MI}-based Approach}
\label{par:ild:approaches:mi}
The \ac{MI}-based \ac{ILD} approach is based on the condition that \ac{IL} occurs in the system if \ac{MI} or \ac{LAS} with \llmis must be significantly greater than $0$, i.e., $\delta(\loss_{ll}) \gg 0$, as per \Cref{sec:ilproblem:ilqd}.
To archive this, we propose to apply the \acf{OTT} on $10$ \ac{MI} estimates obtained using \ac{KFCV} on the \llmi and \midpoint techniques on top-$j$th performing pipeline, as well as baseline methods, denoted by vector $\vec{\hat{I}}_j$.
The \ac{OTT} provides a $p$-value representing the probability of observing the sample mean to be around the actual mean ($0$ in our case) \citep{janez2006statistical}.
The null hypothesis $H_0(\vec{\hat{I}}_j \sim 0)$ implies absence of \ac{IL} in a system.

\paragraph{Classification-based Approaches}
\label{sec:ild:approaches:classification}
Our prior work \citet{ild2022pritha} proposes classification-based \ac{ILD} approaches based on the condition that \ac{IL} occurs if the \bp accuracy significantly surpasses that of a \bpmc, i.e., \ac{LAS} is significantly greater than $0$, as per \Cref{sec:ilproblem:ilqd}.

\subparagraph{\acs{PTT}-based Approach}
\Ac{PTT} is used to compare two samples (generated from an underlying population) where the observations in one sample can be paired with observations in the other sample \citep{janez2006statistical}.
The approach tests the condition that \ac{LAS} is significantly greater than $0$ for \ac{IL} to exist in a system, i.e., $\delta(\metric{ACC}) \gg 0$.
The \pttmc approach achieves this by applying \ac{PTT} between the $10$ accuracy estimates from \ac{KFCV} for the $j$-th best performing \ac{AutoML} pipeline ($g$ minimizing \eqref{eq:erisk}, proxy of \bp) and of $g_{p}^{\textnormal{mc}} $ (proxy of \bpmc, c.f. \Cref{sec:miestimation:llmi:mc}), denoted by $\vec{a}_j$ and $\vec{a}_{mc}$, respectively.
We propose using the corrected version of \ac{PTT}, which accounts for the dependency in estimates due to \ac{KFCV}, to provide an unbiased estimate of the $p$-value \citep{nadeau2003inference}.
The $p$-value represents the probability of obtaining our observed mean accuracy difference, assuming the null hypothesis $H_0(\cdot)$ holds, implying that accuracies are drawn from the same distribution or have nearly zero average difference \citep{janez2006statistical}.
The null hypothesis is $H_0(\acc_j = \acc_{mc})$, which implies no \ac{IL} and the alternate hypothesis $H_1(\acc_j \neq \acc_{mc})$ implies presence of \ac{IL} in the system.
However, \ac{PTT} assumes asymptotic behavior and normal distribution of accuracy differences, which can lead to optimistic $p$-values, and using accuracy can misestimate results on imbalanced datasets \citep{curse2018picek}.

\subparagraph{\acs{FET}-based Approach}
\ac{FET} is a non-parametric test used to determine the probability of independence (or non-dependence) between two classification methods, in this case, classifying instances based on ground truths $\vy$ and classifying them based on \ac{AutoML} predictions $\vhy$ \citep{interpretation1922fisher}.
The \acs{FET}-based approach addresses the class imbalance issue and improves $p$-value estimation by applying the \ac{FET} on $K$ \acp{CM} $\cM_j = \set{\CM{j}{k}}_{k=1}^K$ obtained using \ac{KFCV} to detect \ac{IL}.
If a strong correlation exists between inputs $\vec{x}$ and outputs $y$ in $\cD$, the \ac{AutoML} pipeline's prediction $\hat{y} = g(\vec{x})$ encapsulates input information, and \ac{FET} can asses \ac{IL} using the \ac{CM} (analyzing \ac{LAS} $\delta(\metric{CM})$), which contains relevant information for predicting correct outputs ($\tp, \tn$).
The $p$-value represents the probability of independence between the ground truth $\vy$ and \ac{AutoML} predictions $\vhy$ defined by the null hypothesis ($H_0(\cdot)$).
The null hypothesis $H_0(\prob(\givens{\vy, \vhy}{\CM{}{}}) = \prob(\givens{\vy}{\CM{}{}}) \prob(\givens{\vhy}{\CM{}{}}))$ posits that $\vy$ and $\vhy$ are independent, indicating no \ac{IL}, while the alternative hypothesis 
$H_1(\prob(\given{\vy, \vhy}{\CM{}{}}) \neq \prob(\given{\vy}{\CM{}{}}) \prob(\given{\vhy}{\CM{}{}}))$ suggests significant dependence, implying \ac{IL}.

The \ac{FET} implicitly accounts for dataset imbalance by testing \ac{LAS} using \ac{MCC} (i.e., $\delta(\metric{MCC}) \gg 0$). 
The \ac{MCC} metric, which equally penalizes $\fp$ and $\fn$~\citep{Chicco2021}, is directly proportional to the square root of the $\chi^2$ statistic, i.e., $\card{\metric{MCC}} = \sqrt{\sfrac{\chi^2}{N}}$~\citep{interpretation1922fisher}. 
Since the $\chi^2$ test is asymptotically equivalent to \ac{FET}, this supports using \ac{FET} to evaluate the learnability of $g$ (a proxy for the \bp) via the \ac{CM}, while accounting for class imbalance and yielding accurate $p$-values for detecting \ac{IL}.
We aggregate the $10$ $p$-values obtained using \ac{FET} on $K$ \acp{CM} through \emph{median} and \emph{mean}, referred to as the \fetmedian and \fetmean approaches~\citep{bhaskar2002median}.

\subsubsection{Detection Process}
\label{sec:ild:process:detection}
\begin{figure}[!htb]
	\begin{subfigure}[b]{\textwidth}
		\centering
		\includegraphics[width=\textwidth]{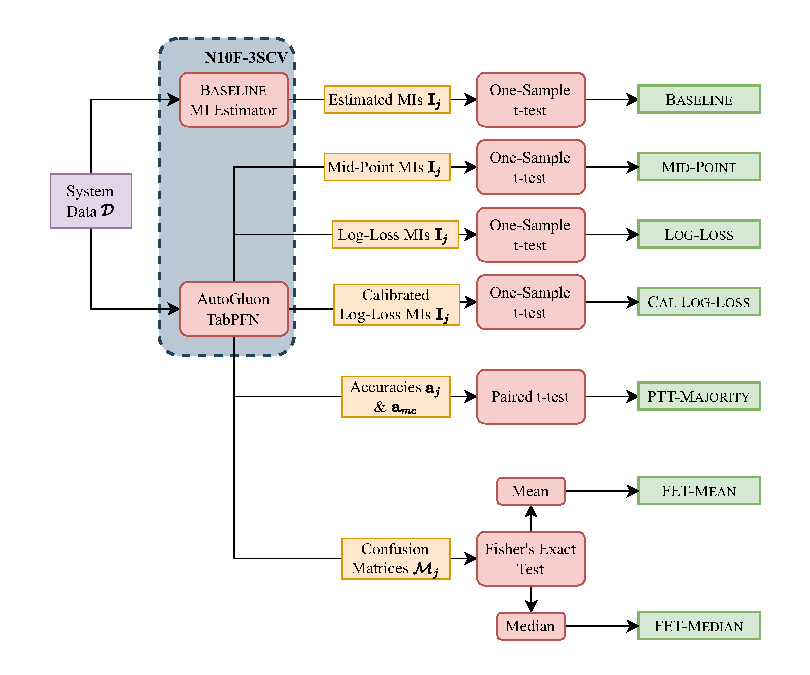}
		\caption{Calculation of $p$-value by different \ac{ILD} approaches}
		\label{subfig:ild_approaches}
	\end{subfigure}
	\begin{subfigure}[b]{\textwidth}
		\centering
		\includegraphics[width=\textwidth]{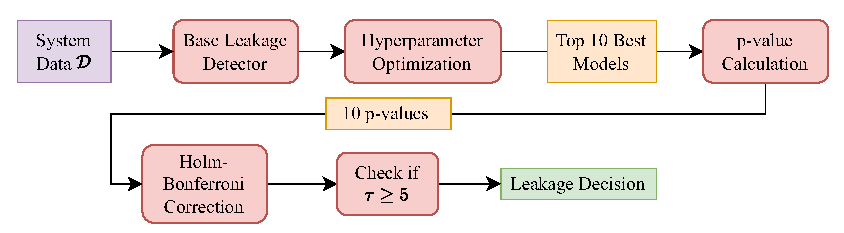}
		\caption{\Acf{IL} detection process}
		\label{subfig:ild_process}
	\end{subfigure}
	\caption{Procedure of using \Ac{ILD} approaches to detect \acp{IL} in a system generating $\cD$}
	\label{fig:ild_combined}
\end{figure}

The \ac{ILD} process involves rigorous estimation, statistical assessment, and correction techniques to detect \ac{IL} in a system confidently.
The process starts with obtaining accurate \ac{MI} or \bp performance estimates using nested \acf{KFCV} with \ac{HPO} and specific parameter ranges provided in \Cref{tab:ranges}.
The dataset $\cD$ is split into \SI{90}{\percent} training and \SI{10}{\percent} test sets using \ac{KFCV} ($10$), conducting \ac{HPO} with $100$ evaluations using \ac{MCCV} with $3$ splits, reserving \SI{30}{\percent} of training data for validation, denoted as \enquote{N$10$F-$3$SCV}, as depicted in \Cref{subfig:ild_approaches}.
Objective functions for \ac{HPO} include \ac{BER} for \pcsoftmax, \autogluon, and \tabpfn, \ac{AIC} for \ac{GMM}, and \ac{MSE} for \ac{MINE} with parameter ranges provided in \Cref{tab:ranges}.
We identify the top-$10$ best-performing pipelines from running \autogluon for $1800$ seconds and top-performing models using \ac{HPO} on \ac{GMM}, \ac{MINE} \pcsoftmax and \tabpfn, using validation loss or accuracy. 
\ac{KFCV} is applied to the top-$10$ models or pipelines, generating $10$ estimates from the entire dataset $\cD$, which statistical tests use to produce $p$-values, which are assessed to detect \ac{IL}.
We obtain $10$ \ac{MI} estimates, accuracies, and \acp{CM} for each of the $j$-th best-performing model/\ac{AutoML} pipeline, denoted by $\vec{\hat{I}}_j$, $\vec{a}_j$ and $\cM_j = \set{\CM{j}{k}}_{k=1}^K$, respectively.
The \ac{OTT} is used on \ac{MI} estimates ($\vec{\hat{I}}_j$), \ac{PTT} compares the accuracies ($\vec{a}_j$) with that of $g_{p}^{\textnormal{mc}} $ ($\vec{a}_{mc}$), and \ac{FET} is applied to \acp{CM} ($\cM_j = \set{\CM{j}{k}}_{k=1}^K$), producing $10$ $p$-values which are aggregated using \emph{mean} and \emph{median} operators.

To enhance \textbf{robustness and reliability} in \ac{ILD}, we use the Holm-Bonferroni correction to ensure accuracy and reliability in our \ac{ILD} framework, which mitigates the influence of overfitting and noisy estimates that can arise when using just one pipeline (model) \citep{holm1979simple}. 
The Holm–Bonferroni method controls the family-wise error rate, minimizing false positives (type-1 errors) by adjusting the rejection criteria $\alpha$ for each hypothesis within our family of null hypotheses, $\cF = \set{H_1, \ldots, H_J}$, ensuring that the significance level of $\cF$ not exceeding predefined threshold of $\alpha = 0.01$ \citep{holm1979simple}.
After obtaining $10$ $p$-values from top-$10$ models/pipelines, we apply the correction to acquire the number of rejected hypotheses or the cut-off parameter $\holm = \card{\cF_r}$, quantifying \ac{IL} detection confidence.
To detect \ac{IL} efficiently, it is imperative to set an appropriate \textbf{rejection threshold} on the cut-off parameter $\holm$.
A higher rejection threshold would help avoid false positives and prevent the detection of non-existent \ac{IL} while decreasing it would avoid missing \acp{IL} occurrences and reduce false negatives.
Based on prior work \citep{ild2022pritha}, setting the rejection threshold to $\floor{\sfrac{J}{2}}$ on the cut-off parameter $\holm$ ensures robust and accurate \ac{IL} detection, which we also use for this study, i.e., $\holm \geq 5$.
This systematic and rigorous approach detects \ac{IL} with a high degree of confidence, ensuring the robustness of our \ac{ILD} framework.

\subsection{Empirical Evaluation}
\label{sec:ild:experiments}
This section outlines the evaluation process for our \ac{ILD} approaches compared to baselines in detecting timing side-channel leaks in OpenSSL TLS servers, as illustrated in \Cref{fig:ild_combined}.
Our main objective is to assess the generalization capability of various \ac{ILD} approaches with respect to the \ac{LAS} of systems, generating both balanced and imbalanced datasets, as outlined in \Cref{tab:openssloverview}.
The overall performance results for selected approaches are discussed in \Cref{sec:ild:results}, with detailed analysis in \ref{asec:generalization:ild}.

\subsubsection{OpenSSL Timing Datasets}
\label{sec:ild:experiments:dataset}
We target side-channel vulnerabilities in cryptographic software to validate our approaches using network traffic generated by a modified OpenSSL TLS server.
Bleichenbacher's attack exploits server behavior to differentiate correctly formatted decrypted messages from incorrect ones \citep{chosen1998bleichenbacher}.
A secure OpenSSL TLS server exhibits no time difference in processing correctly and incorrectly formatted messages, but a vulnerable server does, exposing \ac{IL} through processing time differences.
For our experiments, the timing side-channel or time delay, i.e., observable differences in server computation times when processing messages with correct and manipulated padding, was introduced as per the Java TLS implementation vulnerability \textttnew{CVE-2014-0411}\footnote{\url{https://cve.mitre.org/cgi-bin/cvename.cgi?name=CVE-2014-0411}} \citep{revisiting2014christopher}. 
Datasets were provided by \citet{ild2022funke} from the vulnerable (DamnVulnerableOpenSSL\footnote{\url{https://github.com/tls-attacker/DamnVulnerableOpenSSL}}) OpenSSL TLS server, which exhibits longer computation times for incorrectly formatted messages with manipulated padding, and the non-vulnerable (OpenSSL 1.0.2l\footnote{\url{https://www.openssl.org/source/old/1.0.2/}}), which shows no time delay (no \ac{IL}).
There are $10$ padding manipulations: five cause longer processing times (simulating \ac{IL}, $z=1$), and five do not (no \ac{IL}, $z=0$).
Each \ildd $\data$ contains $10$ binary-class datasets corresponding to $10$ padding manipulations $\cD$, with label $0$ instances corresponding to correctly formatted messages and positive label ($y=1$) instances corresponding to incorrectly formatted messages. \Cref{tab:openssloverview} details the \ildds generated for various time delay values (in $\mu$ seconds) serving as the system's \ac{LAS}, and class imbalances of $r \in \set{0.1, 0.3, 0.5}$ uploaded on OpenML\footnote{\label{footnote:openml}\url{https://www.openml.org/search?type=study&sort=tasks_included&study_type=task&id=383}}.

\begin{table}[!htb] % Overview of the datasets
	\centering
	\fontsize{10}{10}\selectfont % Set desired font size
	\caption{Overview of the OpenSSL timing \ildds used for the \ac{ILD} experiments}
	\label{tab:openssloverview}
	\sisetup{
		table-figures-decimal = 0,
		table-figures-exponent = 0,
		table-number-alignment = center
	}
	\begin{adjustbox}{height=0.10\textwidth, width=0.95\textwidth,center}
		\begin{tabular}{
				S[table-figures-integer=3]|%
				S[table-figures-integer=3]|%
				S[table-figures-integer=3]|%
				S[table-figures-integer=3]
				S[table-figures-integer=3]
				S[table-figures-integer=3]|
				S[table-figures-integer=3]
				S[table-figures-integer=3]
				S[table-figures-integer=3]
				S[table-figures-integer=3]}
			\toprule
			{\begin{tabular}[c]{@{}c@{}}Time Delay \\ (in $\mu$ seconds)\end{tabular}} & {\#\, Folds} & {Imbalance $r$} & \multicolumn{3}{c|}{\ildd $\data$ configuration} & \multicolumn{4}{c}{Dataset $\cD$ configuration} \\
			\midrule
			& & & {\#\, Systems $\card{\data}$} & {\#\, $z=0$} & {\#\,$z=1$} & {$\card{\cD}$} & {\#\, $y=0$} & {\#\,$y=1$} & {\#\, Features}\\
			\midrule
			{$\set{2^{0}, 2^{1}, \ldots, 2^{8}}$} & {3} & {0.1} & {10} & {5} & {5} & {$[943, 1064]$} & {$[849, 958]$} & {$[94, 106]$} & {124}\\
			{$\set{2^{0}, 2^{1}, \ldots, 2^{8}}$} & {3} & {0.3} & {10} & {5} & {5} & {$[1212, 1368]$} & {$[849, 958]$} & {$[363, 410]$} & {124}\\
			{$\set{2^{0}, 2^{1}, \ldots, 2^{8}}$} & {3} & {0.5} & {10} & {5} & {5} & {$[1725, 1929]$} & {$[849, 958]$} & {$[829, 989]$} & {124}\\
			\midrule
			{\{10, 15, \ldots, 35\}} & {10} & {0.1} & {10} & {5} & {5} & {$[962, 2263]$} & {$[866, 2037]$} & {$[96, 226]$} & {$[124, 154]$}\\
			{\{10, 15, \ldots, 35\}} & {10} & {0.3} & {10} & {5} & {5} & {$[1237, 2910]$} & {$[866, 2037]$} & {$[371, 873]$} & {$[124, 154]$}\\
			{\{10, 15, \ldots, 35\}} & {10} & {0.5} & {10} & {5} & {5} & {$[1721, 4084]$} & {$[866, 2037]$} & {$[826, 2104]$} & {$[124, 154]$}\\
			\bottomrule
		\end{tabular}
	\end{adjustbox}
\end{table}

\subsubsection{Experimental Setup}
\label{sec:ild:experiments:setup}
We apply each \ac{ILD} approach depicted in \Cref{subfig:ild_approaches} to a set of \ildds detailed in \Cref{tab:openssloverview}.
As discussed \ildd ($\data = \set{(\cD_i, z_i)}_{i=1}^{10}$) consists of $10$ systems corresponding to $10$ messages with manipulated padding ($y_i=1$), with five systems containing \ac{IL} ($z=1$) and five not ($z=0$).
We evaluate each \ac{ILD} approach using standard binary classification metrics: \accuracy, \fpr, and \fnr (c.f.~\ref{asec:impdetails:evaluationmetric}), on the predicted \ac{IL} decisions ($\vec{\hat{z}} = (\hat{z}_1, \ldots, \hat{z}_{10})$) and the ground truth vector ($\vec{z} = (z_1, \ldots, z_{10})$).

\subsection{Results: Detection Accuracy}
\label{sec:ild:results}
\begin{figure}[!htb]
	\centering
	\includegraphics[width=\textwidth]{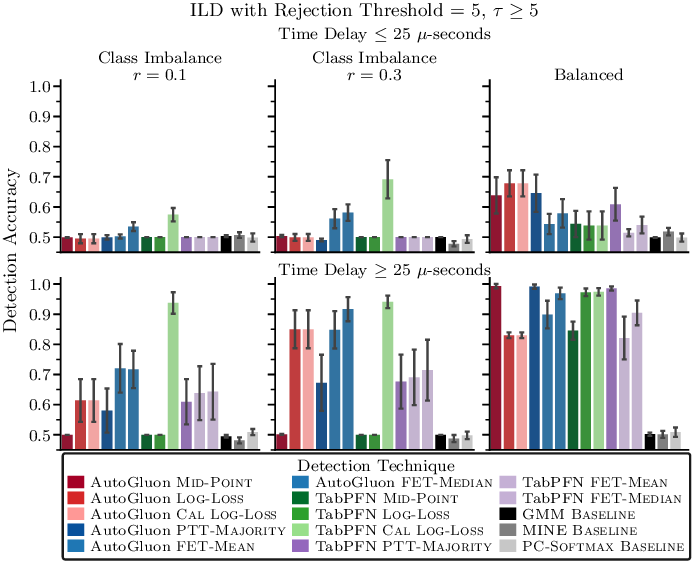}
	\caption{\ac{ILD} performance in detecting OpenSSL TLS timing side-channels}
	\label{fig:ild_overall}
\end{figure}

We evaluate the performance of various \ac{ILD} approaches across different class imbalance levels ($r \in \set{0.1, 0.3, 0.5}$) and time delays.
Our analysis distinguishes between OpenSSL systems with short ($\leq 25$ $\mu$-seconds) and long ($\geq 25$ $\mu$-seconds) delays, reflecting complex and straightforward \ac{IL} cases, respectively.
\Cref{fig:ild_overall} shows the mean detection accuracy and \ac{SE} across \ildds using a rejection threshold of $5$ on $\holm$.
For generalization results across varying \ac{LAS} or time delays, see \ref{asec:generalization:ild}.

\paragraph{Selected \ac{ILD} Approaches}
To identify the best-performing calibration approach for both \tabpfn and \autogluon using \llmi estimation, we assessed their \ac{NMAE} from experiments detailed in \Cref{sec:miestimation:experiments:setup}.
This process determined the optimal calibration approach for each \ac{AutoML} tool separately for balanced ($r = 0.5$) and imbalanced ($r \in {0.1, 0.3}$) datasets.
\autogluon \tscalmi was most effective for both scenarios, while \tabpfn \ircalmi performed best on imbalanced datasets and \tabpfn \pscalmi for balanced ones.
We focus on \fetmedian due to its robustness~\citep{bhaskar2002median}, and prior work~\citep{ild2022pritha} shows it performs similarly to \fetmean at a rejection threshold of $5$ for the cut-off parameter $\holm$.

\subsubsection{Short Time Delay}
Our first set of experiments focuses on systems with short time delays ($\leq 25$ $\mu$-seconds) with more complex or noisy \acp{IL}.

\paragraph{Imbalanced} 
Detecting \acp{IL} with short time delays or \ac{LAS} is challenging. 
Notably, the \tabpfn \calmi approach consistently exhibits high detection accuracy, with \autogluon \fetmedian also performing well. 
For $r = 0.1$, \tabpfn \calmi achieves \SI{69}{\percent} accuracy and \SI{58}{\percent} for $r = 0.3$, while \autogluon \fetmedian detects \SI{54}{\percent} and \SI{58}{\percent} of \acp{IL}, respectively.
However, detecting \acp{IL} in these scenarios remains challenging due to missed \ac{IL} and high \ac{FPR}, as detailed in \ref{asec:generalization:ild}.

\paragraph{Balanced}
Detecting \acp{IL} in systems generating balanced datasets with short time delays remains challenging, but it is more manageable than generating imbalanced ones. 
\autogluon \llmi and \calmi approaches outperform, detecting a significant proportion of \acp{IL} (approximately \SI{68}{\percent}).
\ircalmi enhances \tabpfn \llmi detection accuracy to \SI{67}{\percent}, making it a competent approach, while \pscalmi does not improve its detection performance.
Overall, \autogluon outperforms \tabpfn, with \pttmc detecting over \SI{61}{\percent} of \acp{IL}.

\subsubsection{Long Time Delay}
Our second set of experiments focuses on systems with long time delay ($\geq 25$ $\mu$-seconds) with easily detectable \acp{IL}.

\paragraph{Imbalanced}
Detecting \acp{IL} in systems with larger time delays or \ac{LAS} is more straightforward, with all approaches consistently performing well, detecting more than \SI{50}{\percent} of \acp{IL}.
In particular, the \tabpfn \calmi approach detects approximately \SI{94}{\percent} of \acp{IL}, while \autogluon \fetmedian excels, detecting over \SI{92}{\percent} of \acp{IL} for $r = 0.3$ and approximately \SI{72}{\percent} for $r = 0.1$.
\ac{FET} based approaches perform better than \pttmc in detecting \acp{IL} in imbalanced systems' datasets, also confirmed in \citep{ild2022pritha}.

\paragraph{Balanced}
In systems with balanced datasets and longer time delays, detecting \acp{IL} becomes more straightforward.
\autogluon \midpoint and \tabpfn \ircalmi consistently detect the majority of \acp{IL} (around \SI{99}{\percent}), confirming that \ircalmi provides an effective alternative calibration technique to enhance the performance of the \tabpfn \llmi approach.
In contrast, \tabpfn \pscalmi and \llmi detect approximately \SI{97}{\percent} of \acp{IL}, showing no improvement with the selected calibration method, while \tabpfn \midpoint detects only \SI{85}{\percent}.
Among classification-based approaches, \pttmc consistently outperforms others, detecting \SI{99}{\percent} of \acp{IL}, which contrasts with the findings in \citep{ild2022pritha}.

\subsubsection{Summary}
\label{sec:ild:results:summary}
Detecting \ac{IL} is easier in systems that generate balanced datasets, where \tabpfn \ircalmi, \midpoint, and \pttmc, using \autogluon, show robust performance. 
Contrary to \citet{ild2022pritha}, \pttmc outperforms \ac{FET}-based approaches in balanced datasets, indicating that \ac{ILD} performance varies with \ac{IL} patterns. 
Baseline methods detect only \SI{50}{\percent} of \acp{IL}, reflecting random detection decisions, as discussed in \ref{asec:generalization:ild}.
\tabpfn \calmi consistently outperforms other approaches, highlighting the need for calibrating \tabpfn \llmi for accurate \ac{MI} estimation.
The \midpoint approach is unable to detect \acp{IL} in systems generating imbalanced datasets, as expected. 
For balanced datasets, \ircalmi significantly improves detection accuracy compared to \pscalmi, as shown in \ref{asec:generalization:ild},  indicating that calibration choice is complex based on synthetic dataset results in \Cref{sec:miestimation:results:overall}.
Calibration of \llmi using \autogluon often leads to overfitting and overestimating \ac{MI}, resulting in false positives as shown in \Cref{sec:miestimation:results:overall}.
This discrepancy may result from feature reduction (from $150$ to $[20, 50]$) on an imbalanced dataset using \tabpfn, implying that the choice and necessity of calibration techniques depend on the underlying datasets.

\begin{table}[!htb]
	\centering
	\fontsize{10}{10}\selectfont
	\caption{Runtime comparison (in seconds) for each base learner for all experiments}
	\label{tab:runtimecomparison}
	\sisetup{
		table-number-alignment = center,
		table-figures-integer = 5,
		table-figures-decimal = 2,
		separate-uncertainty = true
	}
    
	\begin{adjustbox}{width=0.85\textwidth,center}
		\begin{tabular}{
				l|
				S[table-figures-integer=5,table-figures-decimal=2]|
				S[table-figures-integer=5,table-figures-decimal=2]}
			\toprule
			{Learner} & {\ac{ILD} Experiments} & {\ac{MI} Estimation Experiments} \\
			\midrule
			\autogluon & 2395.55 \pm 5.21 & 7168.04 \pm 62.77 \\
            \tabpfn & 30513.28 \pm 114.85 & 9970.27 \pm 72.34 \\
			\textsc{GMM Baseline} & 35113.05 \pm 348.72 & 963.79 \pm 22.36 \\
			\textsc{MINE Baseline} & 33447.14 \pm 128.25 & 1962.87 \pm 13.43 \\
			\textsc{PC-softmax Baseline} & 16993.88 \pm 266.83 & 7605.09 \pm 91.36 \\
			Total & 18027.93 \pm 148.12 & 7676.46 \pm 42.60 \\

			\bottomrule
		\end{tabular}
	\end{adjustbox}
\end{table}

\paragraph{Computational Time}
For practical use, \autogluon completes an \ac{ILD} run in approximately $2395$ seconds, while \tabpfn requires around $30513$ seconds, as detailed in \Cref{tab:runtimecomparison}.
Although the transformer-based inference offers more accurate \ac{MI} estimation, making it more suitable when detection accuracy is prioritized over runtime, as shown in \Cref{fig:ild_overall}.
\tabpfn's ability to produce high-quality predictions through a single forward pass reduces the need for calibration in \ac{MI} estimation. 
However, calibration may still be necessary depending on the use of feature reduction and the presence of class imbalance in the dataset.
The baselines are computationally more expensive than \autogluon and perform poorly in detecting \acp{IL}, further indicating that \autogluon offers a favorable trade-off between efficiency and accuracy, making it a computationally effective choice for practical \ac{ILD} deployment.
With \tabpfn now integrated into \autogluon and capable of handling datasets with over $100$ features, it is well-suited for inclusion in a fully automated pipeline that supports appropriate calibration techniques for estimating \ac{MI} via \llmi and enables scalable \ac{ILD} in high-dimensional settings \citep{salinas2024tabrepo}.

\section{Conclusion}
\label{sec:conclusion}
This paper presents a comprehensive framework for \ac{ILD}, leveraging \ite and \slt concepts to enhance cybersecurity by identifying vulnerabilities.
We introduced two techniques for quantifying and detecting \ac{IL} by estimating \ac{MI} between a system's observable and secret information, using the \bp's \llmi and accuracy.
Our methods employ two powerful \ac{AutoML} tools, \tabpfn and \autogluon, for efficient \ac{MI} estimation, offering an automated alternative to traditional statistical techniques or deep \acp{MLP}.
We propose to apply \ac{OTT} to estimated \acp{MI} and enhance robustness and provide confidence in \ac{IL} decisions using Holm-Bonferroni correction, as already established in our prior work \citep{ild2022pritha}.

%advantages
Our empirical results show that our approach, mainly using \tabpfn for calibrated \llmi approximation, effectively estimates \ac{MI} and detects timing side-channel leaks in OpenSSL TLS servers, outperforming state-of-the-art methods. 
The key contributions of our work include a comprehensive \ac{ILD} framework, addressing imbalance with minimal false positives, robustness to noise in generated system datasets, and an adaptable, scalable \ac{IL} detection solution for real-world scenarios.
Furthermore, our work concludes that the choice and requirement of calibration techniques for \llmi estimation depend on the characteristics of the system datasets. 

%future Add references
In future work, we aim to use attention-based transformer embeddings to reduce high-dimensional network traces into compact, semantically meaningful representations, enabling the effective quantification and detection of \ac{IL} while addressing \tabpfn's input limitations.
The new version of \autogluon supports advanced search algorithms such as Hyperband, which can be explored to accelerate the overall process of finding an optimal model using evaluation-efficient successive halving \citep{erickson2022autogluon}.
As \tabpfn is now part of \autogluon’s model pool~\citep{salinas2024tabrepo}, a promising direction is to integrate it with feature selection and dimensionality reduction for building an end-to-end \ac{AutoML} pipeline. This would enable scalable \ac{MI} estimation and \ac{IL} detection on high-dimensional, multi-class datasets ($\nc > 10$), thereby overcoming the limitations of the proposed solution.
Improving the \midpoint approach to account for dataset imbalance and extending the \ac{BER} and \ac{MI} relationship by \citet{ming2013fano} for multiple classes is an essential direction for future work as well.
We plan to extend our experiments to finer-grained imbalance and time-delay variations for deeper robustness analysis. 
Another direction involves integrating permutation-based tests, such as Westfall–Young, and tuning the Holm-Bonferroni rejection threshold to the standard $\alpha = 0.05$ level~\citep{janez2006statistical}.

Additionally, we will explore detecting leaks through side information leaked via channels such as the CPU caches, power consumption, and electromagnetic radiation \citep{picek2023sok, Perianin2021}.
\citep{brillinger2004somedata} showed that the \ac{MSE} of the Bayes-optimal predictor in regression tasks, where the output is real-valued, is directly linked to the \ac{MI} between input and output.
This relationship enables the use of \ac{MI} to quantify and detect leakage in systems vulnerable to cloud-based \acp{SCA}, where continuous-valued user information, such as age or salary, may be exposed \citep{armknecht2017side}. 
Exploring such leakage scenarios represents a promising direction for future work~\citep {zhou2005impact}.
We aim to explore adaptive \ac{IL} detection methods for evolving attack strategies or environments, using approaches such as reinforcement or active learning~\citep{faezi2021ildonline}.

% Acknowledgements should go at the end, before appendices and references


\section*{Acknowledgment}
We are grateful to Dennis Funke, Jan Peter Drees, Karlson Pfannschmidt, Arunselvan Ramaswamy, Björn Haddenhorst, and Stefan Heid for their valuable suggestions. 
We acknowledge computing time on the high-performance computers $\text{Noctua} 2$ at the NHR Center $\text{PC}^2$ for performing the simulations, granted under the project \enquote{hpc-prf-aiafs} (AProSys) project, Förderkennzeichen: 03EI6090E under Abrechnungsobjekt: 3130500154. 
%These are funded by the German Federal Ministry of Education and Research and the state governments participating on the basis of the resolutions of the GWK for the national high-performance computing at universities (www.nhr-verein.de/unsere-partner).

\begin{comment}
	\section*{CRediT authorship contribution statement}
	\textbf{Pritha Gupta}: Conceptualization, Methodology, Validation, Visualization, Formal Analysis, Writing – original draft. 
	\textbf{Marcel Wever}: Supervision, Investigation, Writing– review \& editing. 
	\textbf{Eyke Hüllermeier}: Supervision, Formal Analysis, Writing – review \& editing
	
	\section*{Declaration of competing interest}
	The authors declare that they have no known competing financial interests or personal relationships that could have appeared to influence the work reported in this paper.
	
	\section*{Declaration of generative AI in scientific writing}
	During the preparation of this work Pritha Gupta used Grammarly in order to correct the grammar and shortening the paper
	After using this tool/service, the corresponding author reviewed and edited the content as needed and take(s) full responsibility for the content of the published article.
	
	\section*{Data availability}
	The \ildd datasets used for this research are uploaded on OpenML\textsuperscript{\ref{footnote:openml}}.
\end{comment}

\clearpage
\appendix

\begin{table}[htbp]
	\centering
	\fontsize{8.0}{10}\selectfont % Set desired font size
	\caption{Notation used throughout the paper}
	\label{tab:notations}
	\setlength{\arrayrulewidth}{0.3pt}
	\begin{adjustbox}{width=\textwidth,center}
		\begin{tabular}{p{0.36\textwidth}|p{0.64\textwidth}}
			\toprule
			Symbol   & Meaning         \\
			\midrule
			$[n]$ & Set of integers $\set{1, 2, \dots, n}, n \in \IN$      \\
               $[a,b]$ & Set of integers $\set{a, a+1, \dots, b}, \, a,b \in \IN$      \\
			%$[n]_0$ & Set of integers $\set{0, 1, \dots, n-1}, n-1 \in \IN$      \\
			$\indic{A}$ & Indicator function which is $1$ if statement $A$ is true and $0$ otherwise \\
			$\pjoint(\cdot), \, \hpjoint(\cdot)$ & Actual and predicted joint \ac{PDF} between $(X, Y)$ \\
			$\pjoint(\vec{x}, y) = \pjoint(\vec{x}, y)$ & Joint \ac{PDF} of $X$ and $Y$, at point $(\vec{x}, y)$ \\
			$\pxmarg(\cdot), \, \hpxmarg(\cdot)$ & Actual and predicted marginal \ac{PDF} of $X$ \\
			$\pxmarg(\vec{x}) = \pxmarg(\vec{x})$ & Probability mass of input $\vec{x}$ \\
			$\pymarg(\cdot), \hpymarg(\cdot)$ & Actual and predicted marginal \ac{PMF} of $Y$ \\
			$\pymarg(y) = \pymarg(y)$ & Probability of class label $y$ \\ 
			$\pygivenx(\cdot), \, \hpygivenx(\cdot)$ & Actual and predicted conditional \ac{PDF} of $Y$ given $X$ \\
			$\pygivenx(\given{y}{\vec{x}}) = \pygivenx(\given{y}{\vec{x}})$ & Probability of $y$ given $\vec{x}$\\ 
			$\pxgiveny(\cdot), \, \hpxgiveny(\cdot)$ & Actual and predicted conditional \ac{PDF} of $Y$ given $X$ \\
			$\pxgiveny(\given{\vec{x}}{y}) = \pxgiveny(\given{\vec{x}}{y})$ & Probability of $\vec{x}$ given $y$ \\
			\midrule
			$X$ & Input ($\vec{x} \in \IR^d$) random variable (d-dimensional continuous) \\
			$Y$ & Output ($y \in [\nc]$) random variable (discrete) \\
			$\cX \in \IR^d$ & Input Space, set of $\vec{x}$ sampled from $X$ \\
			$\cY \in [\nc]$ & Output Space, set of $y$ sampled from $Y$ \\
			$\cD = \set{(\vec{x}_i, y_i)}_{i=1}^N$ & Classification dataset \\
			$r$ & Imbalance in a dataset $\cD$ \\
			$\noise $ & Noise in a dataset $\cD$ \\
			\midrule
			$\mi(X; Y)$ & \ac{MI} between $X$ and $Y$ \\
			$\ent(\given{Y}{X})$ & Conditional entropy for $Y$ given $X$ \\ %, i,e uncertainty about $Y$ that remains when we know $X$ \\
			$\ent(X)\, , \ent(Y)$ & Entropy for random variable $X$ and $Y$ \\
			$\ent_2(a) = -(a)\log(a) - (1-a)\log(1-a)$ & Binary cross-entropy function for $a \in (0,1)$ \\
			\midrule
			$\metricbp{Err}{g^{\textnormal{bc}}}, \metricbp{Err}{g^{\textnormal{mc}}}$ & The \berr and \errn of \bpmc\\
			$\delta(\loss_{(\cdot)}) = \loss_{(\cdot)}(g^{\textnormal{mc}}) - \loss_{(\cdot)}(g^{\textnormal{bc}})$ & \multirow{2}{0.64\textwidth}{\Ac{LAS} is the difference in performance of $g^{\textnormal{bc}}$ and $g^{\textnormal{mc}}$ quantifying \ac{IL}} \\
			$\delta(\met_{(\cdot)}) = \card{\met_{(\cdot)}(g^{\textnormal{mc}}) - \met_{(\cdot)}(g^{\textnormal{bc}})}$ &\\
			$\ild\, ,\data $ & \Ac{ILD} function and \ildd\\
			$H_0(\text{condition})\, , H_1(\text{condition})$ & Null and Alternate hypothesis for statistical tests\\
			$\holm \in [J], J \in \IN,$ & Cut-off parameter on $J$ hypothesis for Holm-Bonferroni correction \\
			$\alpha = 0.01$ & Rejection threshold on $H_0$ (accept $H_1$) for statistical tests\\
			\bottomrule
		\end{tabular}
	\end{adjustbox}
\end{table}

\section{Additional Experimental Details}
\label{asec:hporanges}
This section will list all experimental details excluded from the main paper for conciseness.
Hyperparameters and their ranges for \tabpfn, \autogluon, baseline \ac{MI} estimators, and \ac{ILD} approaches are provided in \Cref{tab:ranges}.
We provide a detailed explanation of the algorithms used to generate synthetic system datasets, a description of the performance metrics, and the \ac{AutoML} tools employed in the experiments implemented in the Python package\footnote{\label{footnote:github}\url{https://github.com/LeakDetectAI/AutoMLQuantILDetect}}, along with the scripts used for running the experiments in \Cref{sec:miestimation,sec:ild:experiments} and additional results\footnote{\label{footnote:githubexp}\url{https://github.com/LeakDetectAI/automl-qild-experiments/}}.
The $10$ \ildds per time delay were transformed into multi-class datasets with $11$ padding classes and uploaded to OpenML\textsuperscript{\ref{footnote:openml}}.

\begin{sidewaystable}[htbp]
	\centering
	\caption{Hyperparameter ranges for \ac{AutoML} tools: \autogluon models and \tabpfn including the \ac{MI} estimation baseline approaches (\ac{GMM}, \Ac{MINE}, and \pcsoftmax)}
	\label{tab:ranges}
	\begin{adjustbox}{width=\textwidth, center}
		\begin{tabular}{
				l|
				S[table-figures-integer=1]|
				S[table-figures-integer=1]|
				S[table-figures-integer=1]|
				S[table-figures-integer=1]|
				S[table-figures-integer=1]|
				S[table-figures-integer=1]|
				S[table-figures-integer=1]|
				S[table-figures-integer=1]}
			\toprule
			\multicolumn{9}{c}{\textbf{\autogluon}} \\
			\midrule
			\multicolumn{9}{c}{Tree-based Ensemble Models} \\
			\midrule
			{Learner}    & {Learning Rate} & {\#\, Estimators} & {Max Depth} & {\#\, Leaves} & {Feature Fraction} & {Bagging Fraction} & {Min Data in Leaf} & {Lambda L1 / Lambda L2} \\
			\midrule
			\Ac{LightGBM}    & {[0.01, 0.5]} & {[20, 300]} & {[3, 20]} & {[20, 300]} & {[0.2, 0.95]} & {[0.2, 0.95]} & {[20, 5000]} & {[$\expnumber{1}{-6}$, $\expnumber{1}{-2}$]} \\
			\Ac{CatBoost}    & {[0.01, 0.5]} & NA & {[4, 10]} & NA & NA & NA & NA & {[0.1, 10]} \\
			\Ac{XGBoost}    & {[0.01, 0.5]} & {[20, 300]} & {[3, 10]} & NA & NA & NA & NA & NA \\
			\Ac{RF}    & NA & {[20, 300]} & {[6, 20]} & NA & NA & NA & NA & NA \\
			\Ac{XT}    & NA & {[20, 300]} & {[6, 20]} & NA & NA & NA & NA & NA \\
			\midrule
			\multicolumn{9}{c}{Neural Networks (\Acp{MLP})} \\
			\midrule
			{}    & {Learning Rate} & {Dropout Prob} & {\#\, Layers} & {\#\, Units} & \multicolumn{4}{c}{Other Parameters} \\
			\midrule
			FASTAI   & {[$\expnumber{1}{-5}$, $\expnumber{1}{-1}$]} & {[0.0, 0.5]} & NA & NA & NA & NA & NA & NA \\
			NN\_TORCH  & {[$\expnumber{1}{-5}$, $\expnumber{1}{-1}$]} & {[0.0, 0.5]} & {[2, 20]} & {[8, 256]} & NA & NA & NA & NA \\
			%\midrule
			%\multicolumn{9}{c}{Nearest Neighbor} \\
			%\midrule
			%{} & {Weights} & {p} & {\#\, Neighbors} & \multicolumn{5}{c}{Other Parameters} \\
			%\midrule
			%KNN     & NA & NA & {[3, 5]} & NA & NA & NA & NA & NA \\
			\midrule
			\multicolumn{9}{c}{\textbf{\tabpfn}} \\
			\midrule
			{Learner} & {Reduction Technique} & {\#\, Reduced Features} & {\#\, Ensembles} & \multicolumn{5}{c}{Other Parameters} \\
			\midrule
			\tabpfn & {\Ac{RF} , \Ac{XT}}  & {[10, 50]} & {[32, 200]} & NA & NA & NA & NA & NA \\
			\midrule
			\multicolumn{9}{c}{\textbf{Baselines}} \\
			\midrule
			{Learner} & {Reduction Technique} & {\#\, Reduced Features} & {Covariance Matrix Type} & {Regularization Strength} & \multicolumn{4}{c}{Other Parameters} \\
			\midrule
			\ac{GMM} & {\Ac{RF} , \Ac{XT}} & {[10, 50]} & {\set{Full, Diagonal, Tied, Spherical}} & {[$\expnumber{1}{-10}$, $\expnumber{1}{-1}$]} & NA & NA & NA & NA\\
			\midrule
			{}    & {Learning Rate} & {Optimizer Type} & {\#\, Layers} & {\#\, Units} & {Regularization Strength} & {Early Stopping} & {Batch Normalization} & {Other Parameters} \\
			\midrule
			\ac{MINE}  & {[$\expnumber{1}{-5}$, $\expnumber{1}{-1}$]} & {\set{\rmsprop, \sgd, \adam}} & {[1, 50]} & {[2, 256]} & {[$\expnumber{1}{-10}$, 0.2]} & {\set{True, False}} & {\set{True, False}} & NA \\
			\pcsoftmax  & {[$\expnumber{1}{-5}$, $\expnumber{1}{-1}$]} & {\set{\rmsprop, \sgd, \adam}} & {[1, 50]} & {[2, 256]} & {[$\expnumber{1}{-10}$, 0.2]} & {\set{True, False}} & {\set{True, False}} & NA \\
			\bottomrule
		\end{tabular}
	\end{adjustbox}
\end{sidewaystable}

\subsection{Simulating Synthetic Systems}
\label{asec:syntheticsystems}
We use the \ac{MVN} distribution, which is ideal for simulating real-world vulnerable and non-vulnerable systems generating classification datasets, providing a straightforward means of obtaining ground truth \ac{MI}, as described in \ref{sec:miestimation:syntheticsystems:gtmi}.
The \ac{MVN} distribution is widely used for benchmarking classifiers due to its simplicity, ability to model inter-variable correlations, and well-established statistical properties~\cite[chap.~2]{bishop2006pattern}. 
Its use is further supported by the central limit theorem, which justifies normality in aggregated data~\cite[chap.~4]{bishop2006pattern}. We simulate both balanced and imbalanced datasets with varying \ac{LAS} or \ac{MI} levels using \ac{MVN} perturbation and proximity techniques.

\subsubsection{Generation Method}
\label{asec:syntheticsystems:method}
Synthetic datasets are generated using \ac{MVN} to define the joint \ac{PDF} ($\pjoint$) between $X$ and $Y$, inducing their marginals on $X$ and $Y$.
The dataset $\cD$ is created by sampling instances from $\pxgiveny(\cdot)$ with class distribution $\pymarg(y)$, as illustrated in \Cref{fig:synthetic_datasets}.

\paragraph{Formal Definition of \acp{PDF}}
\label{par:syntheticsystems:method:pdfs}
We define the joint distribution $\pjoint(\cdot)$ and marginal on $Y$ $\pxmarg(\cdot)$, required to generate the dataset $\cD$ by sampling $(\vec{x}_i, y_i) \sim \pxgiveny(\given{\vec{x}_i}{y_i}), \forall i \in [N]$ with class distribution defined by $\pymarg(y)$.
These \acp{PDF} are used to define the conditionals $\pygivenx(\cdot)$ and $\pxgiveny(\cdot)$ and induce the marginal on $\cX$, denoted by $\pxmarg(\cdot)$.
The conditional \ac{PDF} on $X$ given $Y$ is defined for class $\ncs$ as
\begin{align}
\pxgiveny(\given{\vec{x}}{\ncs}) = \mvn(\vec{\mu}_m, \Sigma) \, .
\label{eq:mvnpxgiveny}
\end{align}
where $\Sigma$ is the covariance matrix and $\vec{\mu}_m$ is the mean vector for class $\ncs$.
The covariance matrix should be positive semi-definite and is typically sampled using eigenvalue decomposition \citep{bishop2006pattern}.

\subparagraph{Generating Imbalanced Datasets: Marginal on $Y$}
\label{par:syntheticsystems:method:pdfs:imbalance}
The parameter $r$, referred to as \textbf{class imbalance}, is the minimum proportion of instances for any class $\ncs \in [\nc]$, defined as $r = \min_{\ncs \in [\nc]} \frac{\card{\set{(\vec{x}_i, y_i) \in \cD \; \lvert \; y_i=\ncs}}}{\card{\cD}}$.
For balanced datasets, $r = \sfrac{1}{\nc}$, and for imbalanced ones, $r < \sfrac{1}{\nc}$, i.e., $r \in (0, \sfrac{1}{\nc}]$.
To generate imbalanced multi-class datasets ($\nc>2$), we introduce two methods: \emph{Minority} and \emph{Majority}, with an example class frequency in each case shown in \Cref{fig:imbalanced_datasets}.
In the \emph{Minority} method, the \textbf{minority class} is assigned $r$ fraction of total data points, and remaining samples are uniformly distributed among other classes ($>\frac{N}{\nc}$).
While in the \emph{Majority} method, all classes apart from the selected \textbf{majority class} is assigned $r$ fraction of total data points ($<\frac{N}{\nc}$), and the \textbf{majority class} gets the remaining data points.
The marginal on $Y$ for a balanced dataset with $r = \sfrac{1}{\nc}$ is defined as $\pymarg(\ncs) = \sfrac{1}{\nc}$ and for imbalanced datasets using \emph{Minority} or \emph{Majority} \genmethods is defined as
\begin{equation}
\gentype =
\begin{cases}
\text{Majority} & \begin{cases}
\pymarg(\ncs) = r, & \text{if } \ncs \in [\nc] \setminus \nc \\
\pymarg(\ncs) = 1 - r \cdot (\nc - 1), & \text{if } \ncs = \nc 
\end{cases} \\
\hline

\text{Minority} & \begin{cases}
\pymarg(\ncs) = \frac{1-r}{\nc-1}, & \text{if } \ncs \in [\nc] \setminus \nc \\
\pymarg(\ncs) = r, & \text{if } \ncs = \nc
\end{cases} 
\end{cases}
\label{eq:mvnpymarg}
\end{equation}
Using these, the joint distribution $\pjoint(\vec{x},\ncs)$ for class $\ncs$ is defined as $\pjoint(\vec{x},\ncs) = \pymarg(\ncs) \product \pxgiveny(\given{\vec{x}}{\ncs})$, which induces a marginal on $X$ as
\begin{equation}
\pxmarg(\vec{x}) =\sum_{\ncs=1}^{\nc} \pymarg(\ncs) \product \pxgiveny(\given{\vec{x}}{\ncs}) = \sum_{\ncs=1}^{\nc} \pymarg(\ncs) \product \mvn(\mu_m, \Sigma)
\label{eq:mvnpxmarg}
\end{equation}

\subparagraph{Conditional on $Y$ given $X$}
The conditional $\pygivenx(\cdot)$ on $Y$ given $X$ for instance $(\vec{x}, \ncs)$ is defined as
\begin{align}
\pygivenx(\given{\ncs}{\vec{x}}) &= \frac{\pymarg(\ncs) \product \pxgiveny(\given{\vec{x}}{\ncs})}{\sum_{\ncs=1}^{\nc} \pymarg(\ncs) \product \pxgiveny(\given{\vec{x}}{\ncs})} = \frac{\pjoint(\vec{x},\ncs)}{\pxmarg(\vec{x})} \, . 
\label{eq:mvnpygivenx}
\end{align}

\subsubsection{Introducing Noise ($\epsilon$)}
\label{asec:syntheticsystems:noise}
To simulate real-world \ac{IL} scenarios in cryptographic systems, we introduce noise, which decreases the certainty about output $y_i \in \cY$ with more observed inputs $\vec{x}_i \in \cX$, resulting in $\ent(\given{Y}{X})>0$ and $\mi(X; Y) < \ent(Y)$.
We propose two methods: the \ac{MVN} perturbation and proximity techniques.
The perturbation technique introduces noise by flipping a percentage of outputs or classes in the dataset. 
In contrast, the proximity technique reduces the distance between mean vectors ($\vec{\mu}_{\ncs}$) of each class, leading to overlap between the Gaussians $\mvn(\vec{\mu}_{\ncs}, \Sigma)$.
These methods simulate scenarios where cryptographic systems leak no information (non-vulnerable) with $\mi(X; Y) = 0$. 

\paragraph{\ac{MVN} Perturbation Technique}
\label{par:syntheticsystems:noise:perturbation}
This approach introduces noise by flipping a percentage, denoted by $\epsilon$, of class labels ($y \sim \cY$)  in the dataset, simulating perturbed systems to generate classification datasets.
This technique modifies the conditional \acp{PDF} $\pygivenx(\cdot)$ in \eqref{eq:mvnpygivenx} and  $\pxgiveny(\cdot)$ in \eqref{eq:mvnpxgiveny}.

\subparagraph{Output Variable $Y_{\epsilon}$}
Let random variables $B \sim \text{Bernoulli}(p=\noise, q=1-\noise)$ and $\tilde{Y} \sim \text{Categorical}(\pymarg(1), \ldots, \pymarg(\nc))$ are independent of $\set{X,Y}$, for a fixed $\epsilon$.
The modified output random variable $Y_{\epsilon}$ is defined as $Y_{\epsilon} = Y \product \indic{B=0} + \indic{B=1} \product \tilde{Y}$
The marginal distributions on $X$ remain unchanged because only the class labels are flipped.
Accordingly, the modified marginal on $Y_{\epsilon}$ is derived as $\prob_{Y_{\epsilon}}(Y_{\epsilon} = y) = \epsilon \product \pymarg(Y = y) + (1-\epsilon) \product \pymarg(Y = y) = \pymarg(Y = y)$.
So, the marginal on $Y_{\epsilon}$ is the same as that of $Y$.

\subparagraph{Conditional on $Y_{\epsilon}$ given $X$} 
The conditional on the modified output variable $Y_{\epsilon}$ given $X$ is
\begin{align}
    \prob_{\givens{Y_{\epsilon}}{X}}(\given{Y_{\epsilon}=\ncs}{\vec{x}}) &= \prob_{B}(B=0) \product \pygivenx(\given{\ncs}{\vec{x}}) + \prob_{B}(B=1) \product \prob_{\tilde{Y}}(\tilde{Y} = \ncs) \\
    &= (1-\noise) \product \pygivenx(\given{\ncs}{\vec{x}}) + \noise \product \pymarg(\ncs) \, .
    \label{eq:modified_mvnpygivenx}
\end{align}
\subparagraph{Conditional on $X$ Given $Y_{\epsilon}$}
The conditional $\prob_{\givens{X}{Y_{\epsilon}}}(\given{\vec{x}}{Y_{\epsilon}=\ncs})$ combines multiple \ac{MVN} components is defined as
\begin{equation*}
\prob_{\givens{X}{Y_{\epsilon}}}(\given{\vec{x}}{Y_{\epsilon}=\ncs}) = \frac{\prob_{\givens{Y_{\epsilon}}{X}}(\given{Y_{\epsilon}=\ncs}{\vec{x}}) \pxmarg(\vec{x})}{\pymarg(\ncs)} , .
\end{equation*}
This results in a more complex distribution, which is more difficult to approximate for performing \ac{MI} estimation.  

\paragraph{\ac{MVN} Proximity Technique}
\label{par:syntheticsystems:noise:proximity}
Our second approach introduces noise in the simulated systems by reducing the distance between the Gaussians generated by \ac{MVN} distributions.
This is achieved by moving the mean vectors $\vec{\mu}'_{\ncs}, \forall \ncs \in [\nc]_0$ corresponding to \ac{MVN} distribution representing each class closer to each other.
The updated \acp{MVN} for each class $\ncs$ is defined by $\mvn(\vec{\mu}'_{\ncs}, \Sigma)$, which in turn represents the conditional \ac{PDF} $\pxgiveny(\cdot)$ for the underlying generated system dataset.
Notably, when introducing proximity, only the original \ac{MVN} is modified, which only modifies the conditional on $X$ given $Y$, i.e., $\pxgiveny'(\given{\vec{x}}{\ncs}) = \mvn(\vec{\mu}'_{\ncs}, \Sigma)$. 
Consequently, the conditional \ac{PDF} $\pygivenx(\cdot)$ and the marginal on $X$ can be computed using \eqref{eq:mvnpygivenx} and \eqref{eq:mvnpxmarg} as
\begin{align}
    \pygivenx'(\given{\ncs}{\vec{x}}) &= \frac{\pymarg(\ncs) \product \pxgiveny'(\given{\vec{x}}{\ncs})}{\sum_{\ncs=1}^{\nc} \pymarg(\ncs) \product \pxgiveny'(\given{\vec{x}}{\ncs})} = \frac{\pjoint'(\vec{x},\ncs)}{\pxmarg(\vec{x})} \, . 
    \label{eq:proximity_mvnpygivenx}
\end{align}
Therefore, the underlying distribution for synthetic datasets generated through the introduction of noise using the proximity approach remains the same as for the datasets using the \ac{MVN} distribution with updated means.

\subsubsection{Ground truth \ac{MI}}
\label{sec:miestimation:syntheticsystems:gtmi}
By plugging in the conditional $\pygivenx(\cdot)$ and the marginal $\pymarg(\cdot)$ defined above in \eqref{eq:mi_y}, the ground truth \ac{MI} for the generated system dataset $\cD = \set{\vec{x}_i, y_i}$ is approximated as 
\begin{equation}
GI(\cD) \approxeq \frac{1}{N}\sum_{i=1}^{N} \pygivenx(\given{y_i}{\vec{x}_i}) \log \left(\pygivenx(\given{y_i}{\vec{x}_i})\right) -\sum_{\ncs=1}^{\nc} \pymarg(\ncs) \log \left(\pymarg(\ncs)\right) \, .
\label{eq:gtmvn_mi} 
\end{equation}
For systems simulated using perturbation and proximity techniques, the modified conditional $\prob_{\givens{Y_{\epsilon}}{X}}(\cdot)$ in \eqref{eq:modified_mvnpygivenx} and $\pygivenx'(\given{\ncs}{\vec{x}})$ in \eqref{eq:proximity_mvnpygivenx} and unchanged marginal on $Y$ is used to calculate the ground truth \ac{MI}.

\subsection{Implementation Details}
\label{asec:impdetails}
This section details the implementation setup, including automated tools and calibration methods used for accurate \ac{MI} estimation via \llmi.
Hyperparameters and their ranges for \tabpfn, \autogluon, baseline \ac{MI} estimators, and \ac{ILD} approaches are provided in \Cref{tab:ranges}, used for experiments in \Cref{sec:miestimation,sec:ild:experiments}. 
Python code including \ac{MI} estimation, \ac{ILD} methods, synthetic dataset generation, and the OpenML parser, is available in the GitHub repository\textsuperscript{\ref{footnote:github}}, along with the scripts used for running the experiments in another GitHub repository\textsuperscript{\ref{footnote:githubexp}}.
The 10 \ildds per time delay were transformed into multi-class datasets with 11 padding classes and uploaded to OpenML\textsuperscript{\ref{footnote:openml}}.

\subsubsection{Performance Evaluation Metrics}
\label{asec:impdetails:evaluationmetric}
\citet{koyejo2015} define evaluation measures used for evaluating the performance of classifiers, using the ground truth labels denoted by $\vy = (y_1, \ldots, y_N)$ for a given $\cD = \set{(\vec{x}_i, y_i)}_{i=1}^N$ and the predictions denoted by the vector $\vhy = (\hy_1, \ldots, \hy_N)$, acquired from the \emph{empirical risk minimizers} $g$ and $g_p$ (c.f. \Cref{sec:fundamentals:classification}).

\subparagraph{\accuracy and \err} The accuracy is defined as the proportion of correct predictions, while error-rate is the proportion of incorrect predictions 
\begin{align*} 
\metric{Acc}(\vy, \vhy) \coloneq \dfrac{1}{N}\sum_{i=1}^N\indic{y_i = \hy_i} \,, \quad  
\metric{Err}(\vy, \vhy) \coloneq \dfrac{1}{N}\sum_{i=1}^N\indic{y_i \neq \hy_i} \, .
\end{align*}

\subparagraph{\Acf{CM}}
The \ac{CM} is defined using the \emph{true positive} ($\tp$), \emph{true negative} ($\tn$), \emph{false positive} ($\fp$), and \emph{false negative} ($\fn$) as
\begin{equation*}
\metric{CM}(\vy, \vhy) =
\left[
\begin{array}{cc}
\tn(\vy, \vhy) = \sum\limits_{i=1}^N \indic{y_i=0, \hy_i=0} & 
\fp(\vy, \vhy) = \sum\limits_{i=1}^N \indic{y_i=0, \hy_i=1} \\
\fn(\vy, \vhy) = \sum\limits_{i=1}^N \indic{y_i=1, \hy_i=0} & 
\tp(\vy, \vhy) = \sum\limits_{i=1}^N \indic{y_i=1, \hy_i=1}
\end{array}
\right]
\end{equation*}

\subparagraph{\Ac{FNR} and \Ac{FPR}} The \Ac{FNR} (type 2 error) is defined as the ratio of \emph{false negatives} to the total number of positive instances, while the \Ac{FPR} (type 1 error) is the ratio of \emph{false positives} to the number of negative instances, defined as
\begin{equation*} 
\metric{FNR}(\vy, \vhy) = \tfrac{\fn(\vy, \vhy)}{\fn(\vy, \vhy)+\tp(\vy, \vhy)} \,, \quad \metric{FPR}(\vy, \vhy) = \tfrac{\fp(\vy, \vhy)}{\fp(\vy, \vhy)+\tn(\vy, \vhy)} \, .
\end{equation*}

\subsubsection{AutoML tools}
\label{asec:automl}
In the last decade, the field of \ac{AutoML} has emerged to address the growing need for automating machine learning pipelines, particularly in the absence of expert intervention. While \ac{AutoML} aims to streamline the entire data science workflow, much of the research has focused on the challenge of the \ac{CASH} problem.

Since then, various systems have been devised, demonstrating promising performances for tailoring the choice of \ac{ML} algorithms and the setting of their hyperparameters to a given task, typically comprising a dataset and a loss function \citep{DBLP:journals/jair/ZollerH21}. 
While various \ac{AutoML} systems with complementary strengths have been proposed in the literature, a recent benchmark study \citep{automlbenchmark} suggests \autogluon \citep{erickson2022autogluon} as the \ac{AutoML} system with the best performance across datasets and different tasks.

In \citet{hollmann2023tabpfn}, instead of tackling the \ac{CASH} problem, a general predictor called \tabpfn is fitted across various datasets and can immediately return highly accurate predictions. 
Yielding competitive performance to \autogluon, \tabpfn suggests itself as a state-of-the-art \ac{AutoML} tool that returns results substantially faster than other \ac{AutoML} systems.
Therefore, we selected \autogluon and \tabpfn to approximate the Bayes predictor in our methods for estimating \ac{MI} and detecting \acp{IL} in systems \citep{erickson2022autogluon, hollmann2023tabpfn}.

\paragraph{\autogluon} 
This \ac{AutoML} tool allows the customization of the set of considered \ac{ML} methods, including \ac{GBM} tree-based models and deep \acp{NN}. 
For our study, we limited the search space of \autogluon to consistent classifiers, i.e., tree-based ensemble classifiers like \ac{RF}, \ac{XT}, \ac{GBM} algorithms (\ac{LightGBM}, \ac{CatBoost}, \ac{XGBoost}), and \ac{MLP} implemented using \pytorch and \fastai \citep{Biau2008, Mielniczuk1993}.
The search space \autogluon, consisting of learning algorithms with their corresponding hyperparameter ranges, is listed in \Cref{tab:ranges}.

\paragraph{\tabpfn} 
This \ac{AutoML} tool supports up to \num{100} numeric features, so we apply dimensionality reduction, using \ac{RF} or \ac{XT} when $d > 100$.
We fine-tune \tabpfn using \ac{HPO}, optimizing parameters such as the number of reduced features, reduction method, and number of prior-fitted models, with hyperparameter ranges listed in \Cref{tab:ranges}.

\section{Generalization Capability of \ac{ILD} Approaches}
\label{asec:generalization:ild}
\begin{figure}[!htb]
	\centering
	\includegraphics[width=\textwidth]{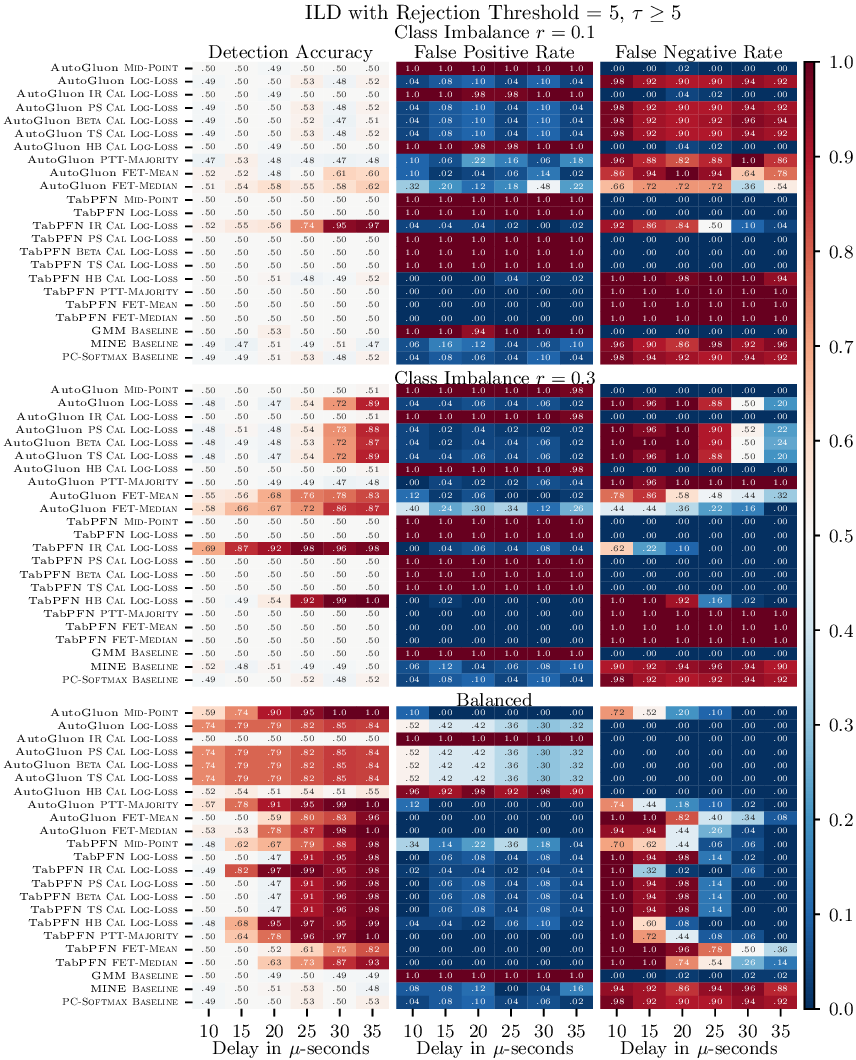}
	\caption{Performance of \ac{ILD} approaches versus time delay with \SI{5}{\micro\second} step}
	\label{fig:timing_threshold5_stepof5}
\end{figure}

\begin{figure}[!htb]
	\centering
	\includegraphics[width=\textwidth]{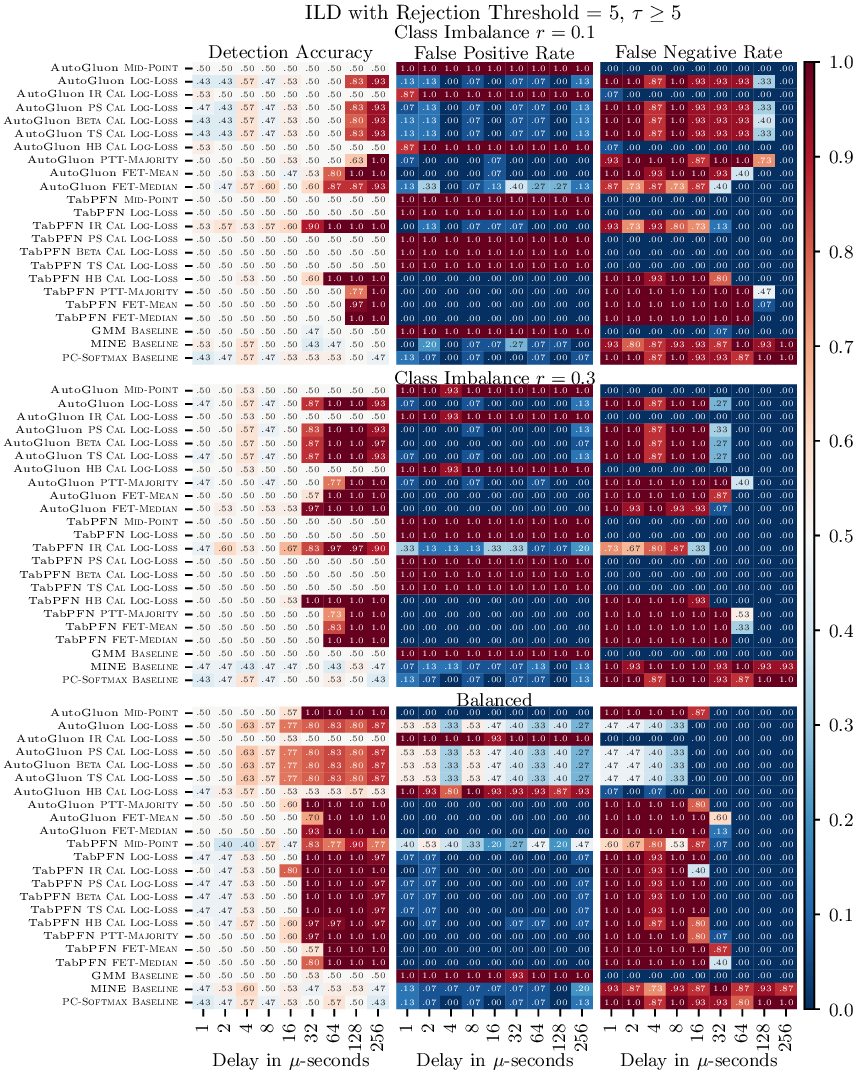}
	\caption{Performance of \ac{ILD} approaches versus time delay with \SI{2}{\micro\second} logarithmic step}
	\label{fig:timing_threshold5_log2}
\end{figure}

This section evaluates the generalization performance of our \ac{ILD} approaches against state-of-the-art baselines. 
We assess \ac{ILD} methods with respect to time delay—the computation time difference between correct and incorrect messages—capturing the system's \ac{LAS}.
We focus on detection accuracy, \fpr, and \fnr for identifying timing side-channels targeting Bleichenbacher-style attacks.
Heatmaps with a fixed rejection threshold ($\holm=5$) are used to visualize results: \Cref{fig:timing_threshold5_stepof5} uses linear \SI{5}{\micro\second} increments, while \Cref{fig:timing_threshold5_log2} uses logarithmic \SI{2}{\micro\second} steps.

\subsection{\tabpfn}
\label{asec:generalization:ild:tabpfn}
This section analyzes \ac{MI} and classification-based \ac{ILD} approaches using \tabpfn. 
In summary, \tabpfn \ircalmi generalizes well, detecting \SI{60}{\percent} of \acp{IL} at minimal delays of \SI{20}{\micro\second}.
It performs even better on balanced datasets, achieving up to \SI{80}{\percent} accuracy.

\paragraph{Classification-based Approaches}
Detection accuracy drops sharply for all approaches when time delays fall below a critical threshold (\SI{35}{\micro\second} for imbalanced dataset and \SI{16}{\micro\second} for balanced dataset), with all methods failing to detect \acp{IL} and yielding a \SI{100}{\percent} \fnr.
\fetmedian performs best on imbalanced datasets, while \pttmc leads on balanced ones.

\paragraph{\ac{MI}-based Approaches}
Overall, \ircalmi consistently outperforms other approaches across both balanced and imbalanced datasets, with \hbcalmi showing competitive results.
While most approaches overfit and overestimate \ac{MI} on imbalanced data, \ircalmi and \hbcalmi maintain reliable detection and avoid false positives. 
Both methods generalize well, detecting more than \SI{80}{\percent} of \acp{IL}, even under minimal time delays of $\SI{20}{\micro\second}$, specifically for the balanced datasets, underscoring \ircalmi’s strength in enhancing \llmi.

\subsection{\autogluon}
\label{asec:generalization:ild:autogluon}
This section analyzes \ac{MI} and classification-based \ac{ILD} approaches using \autogluon. 
Generally, they detect non-existent \acp{IL} in systems generating imbalanced datasets, especially with time delays under \SI{20}{\micro\second}.
However, in systems generating balanced datasets, they occasionally detect over \SI{70}{\percent} of \acp{IL} with similar short delays.

\paragraph{Classification-based Approaches}
Overall, \ac{FET} based approaches show almost comparable detection performance in systems generating balanced and imbalanced datasets, while \pttmc performs better in the case of balanced systems datasets.
All approaches reliably detect \acp{IL} for time-delays above a critical threshold, with all methods failing to detect \acp{IL} and yielding a \SI{100}{\percent} \fnr, below this threshold (\SI{30}{\micro\second} for imbalanced dataset and \SI{16}{\micro\second} for balanced dataset)).


%Mirroring the findings from \tabpfn, the \pttmc approach using \autogluon consistently stands out as the top performer for balanced systems' datasets.

\paragraph{\ac{MI}-based Approaches}
All methods reliably detect \acp{IL} above a critical time-delay threshold, but fail below it—yielding a \SI{100}{\percent} \fnr (at \SI{25}{\micro\second} for imbalanced and \SI{10}{\micro\second} for balanced datasets).
The midpoint approach performs best on balanced datasets, while \llmi excels on imbalanced ones.
This suggests that calibration (\calmi) does not improve \llmis-based \ac{MI} estimation; in fact, \ircalmi and \hbcalmi degrade it by overfitting and overestimating \ac{MI}, resulting in false positives and a \SI{100}{\percent} \fpr—consistent with our findings in \Cref{sec:ild:results,sec:miestimation:results:overall}.

\subsection{Baselines}
\label{asec:generalization:ild:baselines}
The baselines consistently achieve around \SI{50}{\percent} accuracy on both balanced and imbalanced datasets, underscoring the need for specialized \ac{ILD} approaches.
\ac{MINE} and \pcsoftmax frequently miss \acp{IL}, with detection occasionally reaching \SI{60}{\percent}.
\ac{GMM} tends to overestimate \ac{MI}, resulting in false positives and a \SI{100}{\percent} \fpr due to overfitting in high-dimensional settings.

\subsection{Summary}
\label{asec:generalization:ild:summary}
Overall, we conclude that the calibration techniques (\calmi) have mixed effects on \llmi performance across \tabpfn and \autogluon.
While \ircalmi and \hbcalmi often degrade \autogluon \llmi performance due to overfitting, they improve detection with \tabpfn—especially \ircalmi, which consistently achieves the best accuracy on both balanced and imbalanced datasets.
In contrast, with \autogluon, \fetmedian performs best on imbalanced datasets, while \midpoint and \pttmc are most effective for balanced datasets.
These findings, consistent with results in \Cref{sec:ild:experiments,sec:miestimation:results}, highlight the importance of selecting calibration methods based on the dataset and underlying model.
The \midpoint approach excels on balanced datasets but falsely detects \acp{IL} in imbalanced ones, resulting in a \SI{100}{\percent} \fpr, confirming the intuition in \Cref{sec:miestimation:midpointmi}.
Baseline \ac{ILD} approaches remain suboptimal detecting only \SI{50}{\percent} of \acp{IL}, often with \SI{100}{\percent} \fpr or \fnr.

%\bibliographystyle{elsarticle-num-names}
%\bibliography{references.bib}

\begin{thebibliography}{49}
\expandafter\ifx\csname natexlab\endcsname\relax\def\natexlab#1{#1}\fi
\providecommand{\url}[1]{\texttt{#1}}
\providecommand{\href}[2]{#2}
\providecommand{\path}[1]{#1}
\providecommand{\DOIprefix}{doi:}
\providecommand{\ArXivprefix}{arXiv:}
\providecommand{\URLprefix}{URL: }
\providecommand{\Pubmedprefix}{pmid:}
\providecommand{\doi}[1]{\href{http://dx.doi.org/#1}{\path{#1}}}
\providecommand{\Pubmed}[1]{\href{pmid:#1}{\path{#1}}}
\providecommand{\bibinfo}[2]{#2}
\ifx\xfnm\relax \def\xfnm[#1]{\unskip,\space#1}\fi
%Type = Inproceedings
\bibitem[{Kelsey(2002)}]{john2002compression}
\bibinfo{author}{J.~Kelsey},
\newblock \bibinfo{title}{Compression and information leakage of plaintext},
\newblock in: \bibinfo{booktitle}{Fast Software Encryption}, \bibinfo{publisher}{Springer Berlin Heidelberg}, \bibinfo{address}{Berlin, Heidelberg}, \bibinfo{year}{2002}, pp. \bibinfo{pages}{263--276}.
%Type = Article
\bibitem[{Hettwer et~al.(2019)Hettwer, Gehrer, and G\"uneysu}]{hettwer_application_2020}
\bibinfo{author}{B.~Hettwer}, \bibinfo{author}{S.~Gehrer}, \bibinfo{author}{T.~G\"uneysu},
\newblock \bibinfo{title}{Applications of machine learning techniques in side-channel attacks: {A} survey},
\newblock \bibinfo{journal}{Journal of Cryptographic Engineering} \bibinfo{volume}{10} (\bibinfo{year}{2019}) \bibinfo{pages}{135--162}. \DOIprefix\doi{10.1007/s13389-019-00212-8}.
%Type = Book
\bibitem[{Shabtai et~al.(2012)Shabtai, Elovici, and Rokach}]{shabtai2012dl}
\bibinfo{author}{A.~Shabtai}, \bibinfo{author}{Y.~Elovici}, \bibinfo{author}{L.~Rokach}, \bibinfo{title}{A Survey of Data Leakage Detection and Prevention Solutions}, \bibinfo{edition}{1} ed., \bibinfo{publisher}{Springer US}, \bibinfo{address}{New York, NY}, \bibinfo{year}{2012}. \DOIprefix\doi{10.1007/978-1-4614-2053-8}.
%Type = Inproceedings
\bibitem[{Chatzikokolakis et~al.(2010)Chatzikokolakis, Chothia, and Guha}]{chatzikokolakis2010statistical}
\bibinfo{author}{K.~Chatzikokolakis}, \bibinfo{author}{T.~Chothia}, \bibinfo{author}{A.~Guha},
\newblock \bibinfo{title}{Statistical measurement of information leakage},
\newblock in: \bibinfo{booktitle}{Tools and Algorithms for the Construction and Analysis of Systems}, \bibinfo{publisher}{Springer Berlin Heidelberg}, \bibinfo{address}{Berlin, Heidelberg}, \bibinfo{year}{2010}, pp. \bibinfo{pages}{390--404}.
%Type = Inproceedings
\bibitem[{Gao et~al.(2015)Gao, Ver~Steeg, and Galstyan}]{gao2015misurvey}
\bibinfo{author}{S.~Gao}, \bibinfo{author}{G.~Ver~Steeg}, \bibinfo{author}{A.~Galstyan},
\newblock \bibinfo{title}{{Efficient Estimation of Mutual Information for Strongly Dependent Variables}},
\newblock in: \bibinfo{booktitle}{Proceedings of the Eighteenth International Conference on Artificial Intelligence and Statistics}, volume~\bibinfo{volume}{38} of \textit{\bibinfo{series}{Proceedings of Machine Learning Research}}, \bibinfo{publisher}{PMLR}, \bibinfo{address}{San Diego, California, USA}, \bibinfo{year}{2015}, pp. \bibinfo{pages}{277--286}.
%Type = Article
\bibitem[{Moon et~al.(2021)Moon, Sricharan, and Hero}]{moon2021ensemble}
\bibinfo{author}{K.~R. Moon}, \bibinfo{author}{K.~Sricharan}, \bibinfo{author}{A.~O. Hero},
\newblock \bibinfo{title}{Ensemble estimation of generalized mutual information with applications to genomics},
\newblock \bibinfo{journal}{IEEE Transactions on Information Theory} \bibinfo{volume}{67} (\bibinfo{year}{2021}) \bibinfo{pages}{5963--5996}. \DOIprefix\doi{10.1109/TIT.2021.3100108}.
%Type = Article
\bibitem[{Picek et~al.(2023)Picek, Perin, Mariot, Wu, and Batina}]{picek2023sok}
\bibinfo{author}{S.~Picek}, \bibinfo{author}{G.~Perin}, \bibinfo{author}{L.~Mariot}, \bibinfo{author}{L.~Wu}, \bibinfo{author}{L.~Batina},
\newblock \bibinfo{title}{Sok: Deep learning-based physical side-channel analysis},
\newblock \bibinfo{journal}{ACM Computing Surveys} \bibinfo{volume}{55} (\bibinfo{year}{2023}). \DOIprefix\doi{10.1145/3569577}.
%Type = Article
\bibitem[{Moos et~al.(2021)Moos, Wegener, and Moradi}]{dlla2021moos}
\bibinfo{author}{T.~Moos}, \bibinfo{author}{F.~Wegener}, \bibinfo{author}{A.~Moradi},
\newblock \bibinfo{title}{{DL}-{LA:} {Deep} learning leakage assessment},
\newblock \bibinfo{journal}{IACR Transactions on Cryptographic Hardware and Embedded Systems} \bibinfo{volume}{2021} (\bibinfo{year}{2021}) \bibinfo{pages}{552--598}. \DOIprefix\doi{10.46586/tches.v2021.i3.552-598}.
%Type = Article
\bibitem[{Perianin et~al.(2020)Perianin, Carr\'e, Dyseryn, Facon, and Guilley}]{Perianin2021}
\bibinfo{author}{T.~Perianin}, \bibinfo{author}{S.~Carr\'e}, \bibinfo{author}{V.~Dyseryn}, \bibinfo{author}{A.~Facon}, \bibinfo{author}{S.~Guilley},
\newblock \bibinfo{title}{End-to-end automated cache-timing attack driven by machine learning},
\newblock \bibinfo{journal}{Journal of Cryptographic Engineering} \bibinfo{volume}{11} (\bibinfo{year}{2020}) \bibinfo{pages}{135--146}. \DOIprefix\doi{10.1007/s13389-020-00228-5}.
%Type = Incollection
\bibitem[{Cristiani et~al.(2020)Cristiani, Lecomte, and Maurine}]{cristiani2020mineild}
\bibinfo{author}{V.~Cristiani}, \bibinfo{author}{M.~Lecomte}, \bibinfo{author}{P.~Maurine},
\newblock \bibinfo{title}{Leakage assessment through neural estimation of the mutual information},
\newblock in: \bibinfo{booktitle}{Lecture Notes in Computer Science}, \bibinfo{publisher}{Springer International Publishing}, \bibinfo{address}{Berlin, Heidelberg}, \bibinfo{year}{2020}, pp. \bibinfo{pages}{144--162}. \DOIprefix\doi{10.1007/978-3-030-61638-0_9}.
%Type = Article
\bibitem[{Qin and Kim(2019)}]{zhenyue2019rethinking}
\bibinfo{author}{Z.~Qin}, \bibinfo{author}{D.~Kim},
\newblock \bibinfo{title}{Rethinking softmax with cross-entropy: {Neural} network classifier as mutual information estimator},
\newblock \bibinfo{journal}{CoRR} \bibinfo{volume}{abs/1911.10688} (\bibinfo{year}{2019}). \href{http://arxiv.org/abs/1911.10688}{{\tt arXiv:1911.10688}}.
%Type = Article
\bibitem[{Zhang et~al.(2020)Zhang, Zheng, Nan, Hu, and Yu}]{jiajia_novel_2020}
\bibinfo{author}{J.~Zhang}, \bibinfo{author}{M.~Zheng}, \bibinfo{author}{J.~Nan}, \bibinfo{author}{H.~Hu}, \bibinfo{author}{N.~Yu},
\newblock \bibinfo{title}{A novel evaluation metric for deep learning-based side channel analysis and its extended application to imbalanced data},
\newblock \bibinfo{journal}{IACR Transactions on Cryptographic Hardware and Embedded Systems} \bibinfo{volume}{2020} (\bibinfo{year}{2020}) \bibinfo{pages}{73--96}. \DOIprefix\doi{10.46586/tches.v2020.i3.73-96}.
%Type = Article
\bibitem[{Picek et~al.(2018)Picek, Heuser, Jovic, Bhasin, and Regazzoni}]{curse2018picek}
\bibinfo{author}{S.~Picek}, \bibinfo{author}{A.~Heuser}, \bibinfo{author}{A.~Jovic}, \bibinfo{author}{S.~Bhasin}, \bibinfo{author}{F.~Regazzoni},
\newblock \bibinfo{title}{The curse of class imbalance and conflicting metrics with machine learning for side-channel evaluations},
\newblock \bibinfo{journal}{IACR Transactions on Cryptographic Hardware and Embedded Systems} \bibinfo{volume}{2019} (\bibinfo{year}{2018}) \bibinfo{pages}{209--237}. \DOIprefix\doi{10.13154/tches.v2019.i1.209-237}.
%Type = Inproceedings
\bibitem[{Gupta et~al.(2022)Gupta, Ramaswamy, Drees, H\"ullermeier, Priesterjahn, and Jager}]{ild2022pritha}
\bibinfo{author}{P.~Gupta}, \bibinfo{author}{A.~Ramaswamy}, \bibinfo{author}{J.~Drees}, \bibinfo{author}{E.~H\"ullermeier}, \bibinfo{author}{C.~Priesterjahn}, \bibinfo{author}{T.~Jager},
\newblock \bibinfo{title}{Automated information leakage detection: {A} new method combining machine learning and hypothesis testing with an application to side-channel detection in cryptographic protocols},
\newblock in: \bibinfo{booktitle}{Proceedings of the 14th International Conference on Agents and Artificial Intelligence}, \bibinfo{organization}{INSTICC}, \bibinfo{publisher}{SCITEPRESS - Science and Technology Publications}, \bibinfo{address}{Virtual Event}, \bibinfo{year}{2022}, pp. \bibinfo{pages}{152--163}. \DOIprefix\doi{10.5220/0010793000003116}.
%Type = Inbook
\bibitem[{Cover and Thomas(2005)}]{cover2006elements}
\bibinfo{author}{T.~M. Cover}, \bibinfo{author}{J.~A. Thomas}, \bibinfo{title}{Elements of Information Theory}, \bibinfo{publisher}{Wiley}, \bibinfo{year}{2005}, pp. \bibinfo{pages}{13--55}. \DOIprefix\doi{10.1002/047174882x.ch2}.
%Type = Inproceedings
\bibitem[{Vapnik(1991)}]{vapnik1991principles}
\bibinfo{author}{V.~Vapnik},
\newblock \bibinfo{title}{Principles of risk minimization for learning theory},
\newblock in: \bibinfo{booktitle}{Proceedings of the 4th International Conference on Neural Information Processing Systems}, \bibinfo{publisher}{Morgan Kaufmann Publishers Inc.}, \bibinfo{address}{San Francisco, CA, USA}, \bibinfo{year}{1991}, pp. \bibinfo{pages}{831--838}.
%Type = Book
\bibitem[{Bishop(2006)}]{bishop2006pattern}
\bibinfo{author}{C.~M. Bishop}, \bibinfo{title}{Probability Distributions}, \bibinfo{publisher}{Springer New York, NY}, \bibinfo{address}{Springer New York, NY}, \bibinfo{year}{2006}. \DOIprefix\doi{10.5555/1162264}.
%Type = Article
\bibitem[{Gneiting and Raftery(2007)}]{gneiting2007strictly}
\bibinfo{author}{T.~Gneiting}, \bibinfo{author}{A.~E. Raftery},
\newblock \bibinfo{title}{Strictly proper scoring rules, prediction, and estimation},
\newblock \bibinfo{journal}{Journal of the American Statistical Association} \bibinfo{volume}{102} (\bibinfo{year}{2007}) \bibinfo{pages}{359--378}. \DOIprefix\doi{10.1198/016214506000001437}.
%Type = Inbook
\bibitem[{Devroye et~al.(1996)Devroye, Gy\"orfi, and Lugosi}]{devroye2013probabilistic}
\bibinfo{author}{L.~Devroye}, \bibinfo{author}{L.~Gy\"orfi}, \bibinfo{author}{G.~Lugosi}, \bibinfo{title}{The Bayes Error}, volume~\bibinfo{volume}{31}, \bibinfo{publisher}{Springer New York}, \bibinfo{address}{Springer New York, NY}, \bibinfo{year}{1996}, pp. \bibinfo{pages}{9--20}. \DOIprefix\doi{10.1007/978-1-4612-0711-5}.
%Type = Inproceedings
\bibitem[{Belghazi et~al.(2018)Belghazi, Baratin, Rajeshwar, Ozair, Bengio, Courville, and Hjelm}]{belghazi2018mine}
\bibinfo{author}{M.~I. Belghazi}, \bibinfo{author}{A.~Baratin}, \bibinfo{author}{S.~Rajeshwar}, \bibinfo{author}{S.~Ozair}, \bibinfo{author}{Y.~Bengio}, \bibinfo{author}{A.~Courville}, \bibinfo{author}{D.~Hjelm},
\newblock \bibinfo{title}{Mutual information neural estimation},
\newblock in: \bibinfo{booktitle}{Proceedings of the 35th International Conference on Machine Learning}, volume~\bibinfo{volume}{80}, \bibinfo{publisher}{Proceedings of Machine Learning Research}, \bibinfo{address}{Stockholmsm\"assan, Stockholm, Sweden}, \bibinfo{year}{2018}, pp. \bibinfo{pages}{531--540}.
%Type = Article
\bibitem[{Maia~Polo and Vicente(2022)}]{mariapolo2022effective}
\bibinfo{author}{F.~Maia~Polo}, \bibinfo{author}{R.~Vicente},
\newblock \bibinfo{title}{Effective sample size, dimensionality, and generalization in covariate shift adaptation},
\newblock \bibinfo{journal}{Neural Computing and Applications} \bibinfo{volume}{35} (\bibinfo{year}{2022}) \bibinfo{pages}{18187--18199}. \DOIprefix\doi{10.1007/s00521-021-06615-1}.
%Type = Article
\bibitem[{Tebbe and Dwyer(1968)}]{tebbe1968uncertainty}
\bibinfo{author}{D.~Tebbe}, \bibinfo{author}{S.~Dwyer},
\newblock \bibinfo{title}{Uncertainty and the probability of error {(Corresp.)}},
\newblock \bibinfo{journal}{IEEE Transactions on Information Theory} \bibinfo{volume}{14} (\bibinfo{year}{1968}) \bibinfo{pages}{516--518}. \DOIprefix\doi{10.1109/tit.1968.1054135}.
%Type = Article
\bibitem[{Hellman and Raviv(1970)}]{hellman1970probability}
\bibinfo{author}{M.~Hellman}, \bibinfo{author}{J.~Raviv},
\newblock \bibinfo{title}{Probability of error, equivocation, and the chernoff bound},
\newblock \bibinfo{journal}{IEEE Transactions on Information Theory} \bibinfo{volume}{16} (\bibinfo{year}{1970}) \bibinfo{pages}{368--372}. \DOIprefix\doi{10.1109/tit.1970.1054466}.
%Type = Book
\bibitem[{Fano(1961)}]{fano1961transmission}
\bibinfo{author}{R.~M. Fano}, \bibinfo{title}{Transmission of Information: A Statistical Theory of Communications}, \bibinfo{publisher}{The MIT Press}, \bibinfo{address}{Cambridge, MA}, \bibinfo{year}{1961}.
%Type = Article
\bibitem[{Zhao et~al.(2013)Zhao, Edakunni, Pocock, and Brown}]{ming2013fano}
\bibinfo{author}{M.-J. Zhao}, \bibinfo{author}{N.~Edakunni}, \bibinfo{author}{A.~Pocock}, \bibinfo{author}{G.~Brown},
\newblock \bibinfo{title}{Beyond fano's inequality: Bounds on the optimal f-score, ber, and cost-sensitive risk and their implications},
\newblock \bibinfo{journal}{Journal of Machine Learning Research} \bibinfo{volume}{14} (\bibinfo{year}{2013}) \bibinfo{pages}{1033--1090}.
%Type = Inproceedings
\bibitem[{Szegedy et~al.(2016)Szegedy, Vanhoucke, Ioffe, Shlens, and Wojna}]{szegedyls}
\bibinfo{author}{C.~Szegedy}, \bibinfo{author}{V.~Vanhoucke}, \bibinfo{author}{S.~Ioffe}, \bibinfo{author}{J.~Shlens}, \bibinfo{author}{Z.~Wojna},
\newblock \bibinfo{title}{Rethinking the {Inception} architecture for computer vision},
\newblock in: \bibinfo{booktitle}{Proceedings of the {IEEE} Conference on Computer Vision and Pattern Recognition, {CVPR}, Las Vegas, NV, USA, June 27-30}, \bibinfo{publisher}{{IEEE} Computer Society}, \bibinfo{year}{2016}, pp. \bibinfo{pages}{2818--2826}.
%Type = Article
\bibitem[{Silva~Filho et~al.(2023)Silva~Filho, Song, Perello-Nieto, Santos-Rodriguez, Kull, and Flach}]{silva2023calibration}
\bibinfo{author}{T.~Silva~Filho}, \bibinfo{author}{H.~Song}, \bibinfo{author}{M.~Perello-Nieto}, \bibinfo{author}{R.~Santos-Rodriguez}, \bibinfo{author}{M.~Kull}, \bibinfo{author}{P.~Flach},
\newblock \bibinfo{title}{Classifier calibration: A survey on how to assess and improve predicted class probabilities},
\newblock \bibinfo{journal}{Machine Learning} \bibinfo{volume}{112} (\bibinfo{year}{2023}) \bibinfo{pages}{3211--3260}. \DOIprefix\doi{10.1007/s10994-023-06336-7}.
%Type = Inproceedings
\bibitem[{Kario(2024)}]{kario2024marvin}
\bibinfo{author}{H.~Kario},
\newblock \bibinfo{title}{Everlasting {ROBOT}: The marvin attack},
\newblock in: \bibinfo{editor}{G.~Tsudik}, \bibinfo{editor}{M.~Conti}, \bibinfo{editor}{K.~Liang}, \bibinfo{editor}{G.~Smaragdakis} (Eds.), \bibinfo{booktitle}{Computer Security -- {ESORICS} 2023: 28th European Symposium on Research in Computer Security, The Hague, The Netherlands, September 25--29, 2023, Proceedings, Part III}, volume \bibinfo{volume}{14137} of \textit{\bibinfo{series}{Lecture Notes in Computer Science}}, \bibinfo{publisher}{Springer Nature Switzerland}, \bibinfo{address}{The Hague, The Netherlands}, \bibinfo{year}{2024}, pp. \bibinfo{pages}{243--262}. \URLprefix \url{https://doi.org/10.1007/978-3-031-51479-1_13}. \DOIprefix\doi{10.1007/978-3-031-51479-1_13}.
%Type = Article
\bibitem[{Holm(1979)}]{holm1979simple}
\bibinfo{author}{S.~Holm},
\newblock \bibinfo{title}{A simple sequentially rejective multiple test procedure},
\newblock \bibinfo{journal}{Scandinavian Journal of Statistics} \bibinfo{volume}{6} (\bibinfo{year}{1979}) \bibinfo{pages}{65--70}.
%Type = Article
\bibitem[{Dem\v{s}ar(2006)}]{janez2006statistical}
\bibinfo{author}{J.~Dem\v{s}ar},
\newblock \bibinfo{title}{Statistical comparisons of classifiers over multiple data sets},
\newblock \bibinfo{journal}{Journal of Machine Learning Research} \bibinfo{volume}{7} (\bibinfo{year}{2006}) \bibinfo{pages}{1--30}.
%Type = Article
\bibitem[{Nadeau(2003)}]{nadeau2003inference}
\bibinfo{author}{C.~Nadeau},
\newblock \bibinfo{title}{Inference for the generalization error},
\newblock \bibinfo{journal}{Machine Learning} \bibinfo{volume}{52} (\bibinfo{year}{2003}) \bibinfo{pages}{239--281}. \DOIprefix\doi{10.1023/a:1024068626366}.
%Type = Article
\bibitem[{Fisher(1922)}]{interpretation1922fisher}
\bibinfo{author}{R.~A. Fisher},
\newblock \bibinfo{title}{On the interpretation of \ensuremath{\chi} 2 from contingency tables, and the calculation of p},
\newblock \bibinfo{journal}{Journal of the Royal Statistical Society} \bibinfo{volume}{85} (\bibinfo{year}{1922}) \bibinfo{pages}{87}. \DOIprefix\doi{10.2307/2340521}.
%Type = Article
\bibitem[{Chicco et~al.(2021)Chicco, T{\"o}tsch, and Jurman}]{Chicco2021}
\bibinfo{author}{D.~Chicco}, \bibinfo{author}{N.~T{\"o}tsch}, \bibinfo{author}{G.~Jurman},
\newblock \bibinfo{title}{The matthews correlation coefficient (mcc) is more reliable than balanced accuracy, bookmaker informedness, and markedness in two-class confusion matrix evaluation},
\newblock \bibinfo{journal}{BioData Mining} \bibinfo{volume}{14} (\bibinfo{year}{2021}) \bibinfo{pages}{13}. \DOIprefix\doi{10.1186/s13040-021-00244-z}.
%Type = Article
\bibitem[{Bhattacharya and Habtzghi(2002)}]{bhaskar2002median}
\bibinfo{author}{B.~Bhattacharya}, \bibinfo{author}{D.~Habtzghi},
\newblock \bibinfo{title}{Median of {the\ensuremath{<}i\ensuremath{>}p\ensuremath{<}/i\ensuremath{>}Value} under the alternative hypothesis},
\newblock \bibinfo{journal}{The American Statistician} \bibinfo{volume}{56} (\bibinfo{year}{2002}) \bibinfo{pages}{202--206}. \DOIprefix\doi{10.1198/000313002146}.
%Type = Inproceedings
\bibitem[{Bleichenbacher(1998)}]{chosen1998bleichenbacher}
\bibinfo{author}{D.~Bleichenbacher},
\newblock \bibinfo{title}{Chosen ciphertext attacks against protocols based on the rsa encryption standard pkcs {\#}1},
\newblock in: \bibinfo{booktitle}{Advances in Cryptology --- CRYPTO '98}, \bibinfo{publisher}{Springer Berlin Heidelberg}, \bibinfo{address}{Berlin, Heidelberg}, \bibinfo{year}{1998}, pp. \bibinfo{pages}{1--12}.
%Type = Inproceedings
\bibitem[{Meyer et~al.(2014)Meyer, Somorovsky, Weiss, Schwenk, Schinzel, and Tews}]{revisiting2014christopher}
\bibinfo{author}{C.~Meyer}, \bibinfo{author}{J.~Somorovsky}, \bibinfo{author}{E.~Weiss}, \bibinfo{author}{J.~Schwenk}, \bibinfo{author}{S.~Schinzel}, \bibinfo{author}{E.~Tews},
\newblock \bibinfo{title}{Revisiting {SSL/TLS} implementations: {New} bleichenbacher side channels and attacks},
\newblock in: \bibinfo{booktitle}{23rd USENIX Security Symposium (USENIX Security 14)}, \bibinfo{publisher}{USENIX Association}, \bibinfo{address}{San Diego, CA}, \bibinfo{year}{2014}, pp. \bibinfo{pages}{733--748}.
%Type = Misc
\bibitem[{Funke(2022)}]{ild2022funke}
\bibinfo{author}{D.~Funke}, \bibinfo{title}{Pushing the {AutoSCA} tool to picosecond precision: {Improving} timing side channel detection}, \bibinfo{year}{2022}. \DOIprefix\doi{10.13140/RG.2.2.33070.08005}.
%Type = Inproceedings
\bibitem[{Salinas and Erickson(2024)}]{salinas2024tabrepo}
\bibinfo{author}{D.~Salinas}, \bibinfo{author}{N.~Erickson},
\newblock \bibinfo{title}{Tabrepo: A large scale repository of tabular model evaluations and its automl applications},
\newblock in: \bibinfo{editor}{K.~Eggensperger}, \bibinfo{editor}{R.~Garnett}, \bibinfo{editor}{J.~Vanschoren}, \bibinfo{editor}{M.~Lindauer}, \bibinfo{editor}{J.~R. Gardner} (Eds.), \bibinfo{booktitle}{Proceedings of the Third International Conference on Automated Machine Learning}, volume \bibinfo{volume}{256} of \textit{\bibinfo{series}{Proceedings of Machine Learning Research}}, \bibinfo{publisher}{PMLR}, \bibinfo{year}{2024}, pp. \bibinfo{pages}{19/1--30}. \URLprefix \url{https://proceedings.mlr.press/v256/salinas24a.html}.
%Type = Article
\bibitem[{Erickson et~al.(2020)Erickson, Mueller, Shirkov, Zhang, Larroy, Li, and Smola}]{erickson2022autogluon}
\bibinfo{author}{N.~Erickson}, \bibinfo{author}{J.~Mueller}, \bibinfo{author}{A.~Shirkov}, \bibinfo{author}{H.~Zhang}, \bibinfo{author}{P.~Larroy}, \bibinfo{author}{M.~Li}, \bibinfo{author}{A.~Smola},
\newblock \bibinfo{title}{{AutoGluon}-tabular: {Robust} and accurate {AutoML} for structured data},
\newblock \bibinfo{journal}{arXiv preprint arXiv:2003.06505}  (\bibinfo{year}{2020}).
%Type = Article
\bibitem[{Brillinger(2004)}]{brillinger2004somedata}
\bibinfo{author}{D.~R. Brillinger},
\newblock \bibinfo{title}{Some data analyses using mutual information},
\newblock \bibinfo{journal}{Brazilian Journal of Probability and Statistics} \bibinfo{volume}{18} (\bibinfo{year}{2004}) \bibinfo{pages}{163--182}. \URLprefix \url{http://www.jstor.org/stable/43601047}.
%Type = Inproceedings
\bibitem[{Armknecht et~al.(2017)Armknecht, Boyd, Davies, Gjøsteen, and Toorani}]{armknecht2017side}
\bibinfo{author}{F.~Armknecht}, \bibinfo{author}{C.~Boyd}, \bibinfo{author}{G.~T. Davies}, \bibinfo{author}{K.~Gjøsteen}, \bibinfo{author}{M.~Toorani},
\newblock \bibinfo{title}{Side channels in deduplication: Trade-offs between leakage and efficiency},
\newblock in: \bibinfo{booktitle}{Proceedings of the 2017 ACM on Asia Conference on Computer and Communications Security (ASIA CCS '17)}, \bibinfo{publisher}{Association for Computing Machinery}, \bibinfo{address}{New York, NY, USA}, \bibinfo{year}{2017}, pp. \bibinfo{pages}{266--274}. \URLprefix \url{https://doi.org/10.1145/3052973.3053019}. \DOIprefix\doi{10.1145/3052973.3053019}.
%Type = Article
\bibitem[{Zhou and Feng(2005)}]{zhou2005impact}
\bibinfo{author}{Y.~Zhou}, \bibinfo{author}{D.~Feng},
\newblock \bibinfo{title}{Side-channel attacks: Ten years after its publication and the impacts on cryptographic module security testing},
\newblock \bibinfo{journal}{IACR Cryptology ePrint Archive}  (\bibinfo{year}{2005}). \URLprefix \url{http://eprint.iacr.org/2005/388}.
%Type = Article
\bibitem[{Faezi et~al.(2021)Faezi, Yasaei, Barua, and Faruque}]{faezi2021ildonline}
\bibinfo{author}{S.~Faezi}, \bibinfo{author}{R.~Yasaei}, \bibinfo{author}{A.~Barua}, \bibinfo{author}{M.~A.~A. Faruque},
\newblock \bibinfo{title}{Brain-inspired golden chip free hardware trojan detection},
\newblock \bibinfo{journal}{IEEE Transactions on Information Forensics and Security} \bibinfo{volume}{16} (\bibinfo{year}{2021}) \bibinfo{pages}{2697--2708}. \DOIprefix\doi{10.1109/TIFS.2021.3062989}.
%Type = Inproceedings
\bibitem[{Koyejo et~al.(2015)Koyejo, Ravikumar, Natarajan, and Dhillon}]{koyejo2015}
\bibinfo{author}{O.~Koyejo}, \bibinfo{author}{P.~Ravikumar}, \bibinfo{author}{N.~Natarajan}, \bibinfo{author}{I.~S. Dhillon},
\newblock \bibinfo{title}{Consistent multilabel classification},
\newblock in: \bibinfo{booktitle}{Proceedings of the 28th International Conference on Neural Information Processing Systems - Volume 2}, \bibinfo{publisher}{MIT Press}, \bibinfo{address}{Cambridge, MA, USA}, \bibinfo{year}{2015}, pp. \bibinfo{pages}{3321--3329}. \DOIprefix\doi{10.5555/2969442.2969610}.
%Type = Article
\bibitem[{Z{\"{o}}ller and Huber(2021)}]{DBLP:journals/jair/ZollerH21}
\bibinfo{author}{M.~Z{\"{o}}ller}, \bibinfo{author}{M.~F. Huber},
\newblock \bibinfo{title}{Benchmark and survey of automated machine learning frameworks},
\newblock \bibinfo{journal}{Journal of Artificial Intelligence Research} \bibinfo{volume}{70} (\bibinfo{year}{2021}) \bibinfo{pages}{409--472}. \DOIprefix\doi{10.1613/jair.1.11854}.
%Type = Article
\bibitem[{Gijsbers et~al.(2024)Gijsbers, Bueno, Coors, LeDell, Poirier, Thomas, Bischl, and Vanschoren}]{automlbenchmark}
\bibinfo{author}{P.~Gijsbers}, \bibinfo{author}{M.~L.~P. Bueno}, \bibinfo{author}{S.~Coors}, \bibinfo{author}{E.~LeDell}, \bibinfo{author}{S.~Poirier}, \bibinfo{author}{J.~Thomas}, \bibinfo{author}{B.~Bischl}, \bibinfo{author}{J.~Vanschoren},
\newblock \bibinfo{title}{Amlb: an automl benchmark},
\newblock \bibinfo{journal}{Journal of Machine Learning Research} \bibinfo{volume}{25} (\bibinfo{year}{2024}) \bibinfo{pages}{1--65}.
%Type = Inproceedings
\bibitem[{Hollmann et~al.(2023)Hollmann, M\"uller, Eggensperger, and Hutter}]{hollmann2023tabpfn}
\bibinfo{author}{N.~Hollmann}, \bibinfo{author}{S.~M\"uller}, \bibinfo{author}{K.~Eggensperger}, \bibinfo{author}{F.~Hutter},
\newblock \bibinfo{title}{{TabPFN:} {A} transformer that solves small tabular classification problems in a second},
\newblock in: \bibinfo{booktitle}{The Eleventh International Conference on Learning Representations}, \bibinfo{year}{2023}. \URLprefix \url{https://openreview.net/forum?id=cp5PvcI6w8_}.
%Type = Article
\bibitem[{Biau et~al.(2008)Biau, Devroye, and Lugosi}]{Biau2008}
\bibinfo{author}{G.~Biau}, \bibinfo{author}{L.~Devroye}, \bibinfo{author}{G.~Lugosi},
\newblock \bibinfo{title}{Consistency of random forests and other averaging classifiers},
\newblock \bibinfo{journal}{Journal of Machine Learning Research} \bibinfo{volume}{9} (\bibinfo{year}{2008}) \bibinfo{pages}{2015--2033}.
%Type = Article
\bibitem[{Mielniczuk and Tyrcha(1993)}]{Mielniczuk1993}
\bibinfo{author}{J.~Mielniczuk}, \bibinfo{author}{J.~Tyrcha},
\newblock \bibinfo{title}{Consistency of multilayer perceptron regression estimators},
\newblock \bibinfo{journal}{Neural Networks} \bibinfo{volume}{6} (\bibinfo{year}{1993}) \bibinfo{pages}{1019--1022}. \DOIprefix\doi{10.1016/s0893-6080(09)80011-7}.

\end{thebibliography}

\end{document}